\documentclass[final]{cvpr}

\usepackage{times}
\usepackage{epsfig}
\usepackage{graphicx}
\usepackage{amsmath}
\usepackage{amssymb}

\usepackage{bm}
\usepackage{cases}
\usepackage[linesnumbered,ruled,vlined,lined,boxed,commentsnumbered]{algorithm2e}
\usepackage{subfigure}
\usepackage{indentfirst}
\usepackage{multirow}
\usepackage{boldline}
\usepackage{colortbl}

\usepackage{enumitem}
\usepackage{mathrsfs}
\usepackage{algorithmic}

\setitemize{noitemsep,topsep=0pt,parsep=0pt,partopsep=0pt}

\usepackage[table]{xcolor}

\usepackage{pifont}
\newcommand{\cmark}{\ding{51}}%
\usepackage[pagebackref=true,breaklinks=true,colorlinks,bookmarks=false]{hyperref}

\makeatletter\renewcommand\paragraph{\@startsection{paragraph}{4}{\z@}
  {.5em \@plus1ex \@minus.2ex}{-.5em}{\normalfont\normalsize\bfseries}}\makeatother

\newcolumntype{x}[1]{>{\centering\arraybackslash}p{#1pt}}
\newlength\savewidth\newcommand\shline{\noalign{\global\savewidth\arrayrulewidth\global\arrayrulewidth 1pt}\hline\noalign{\global\arrayrulewidth\savewidth}}
\newcommand{\tablestyle}[2]{\setlength{\tabcolsep}{#1}\renewcommand{\arraystretch}{#2}\centering\footnotesize}

\def\cvprPaperID{3876} 
\def\confYear{CVPR 2021}

\begin{document}

\title{Spatiotemporal Contrastive Video Representation Learning}

\author{
Rui Qian\thanks{~The first two authors contributed equally. This work was performed while Rui Qian worked at Google.}~~$^{1,2,3}$\qquad
Tianjian Meng\footnotemark[1]~~$^{1}$\qquad
Boqing Gong$^{1}$\qquad
Ming-Hsuan Yang$^{1}$ \\
Huisheng Wang$^{1}$\qquad
Serge Belongie$^{1,2,3}$\qquad
Yin Cui$^{1}$ \\
\\
$^{1}$Google Research \qquad $^{2}$Cornell University \qquad $^{3}$Cornell Tech}

\maketitle
\thispagestyle{empty}

\begin{abstract}
We present a self-supervised Contrastive Video Representation Learning (CVRL) method to learn spatiotemporal visual representations from unlabeled videos.
Our representations are learned using a contrastive loss, where two augmented clips from the same short video are pulled together in the embedding space, while clips from different videos are pushed away.
We study what makes for good data augmentations for video self-supervised learning and find that both spatial and temporal information are crucial. 
We carefully design data augmentations involving spatial and temporal cues. Concretely, we propose a temporally consistent spatial augmentation method to impose strong spatial augmentations on each frame of the video while maintaining the temporal consistency across frames. We also propose a sampling-based temporal augmentation method to avoid overly enforcing invariance on clips that are distant in time.  
On Kinetics-600, a linear classifier trained on the representations learned by CVRL achieves 70.4\% top-1 accuracy with a 3D-ResNet-50 (R3D-50) backbone, outperforming ImageNet supervised pre-training by 15.7\% and SimCLR unsupervised pre-training by 18.8\% using the same inflated R3D-50.
The performance of CVRL can be further improved to 72.9\% with a larger R3D-152 (2$\times$ filters) backbone, significantly closing the gap between unsupervised and supervised video representation learning. Our code and models will be available at \href{https://github.com/tensorflow/models/tree/master/official/}{https://github.com/tensorflow/models/tree/master/official/}.
\end{abstract}

\begin{figure}[t]
\centering
\includegraphics[width=0.95\columnwidth]{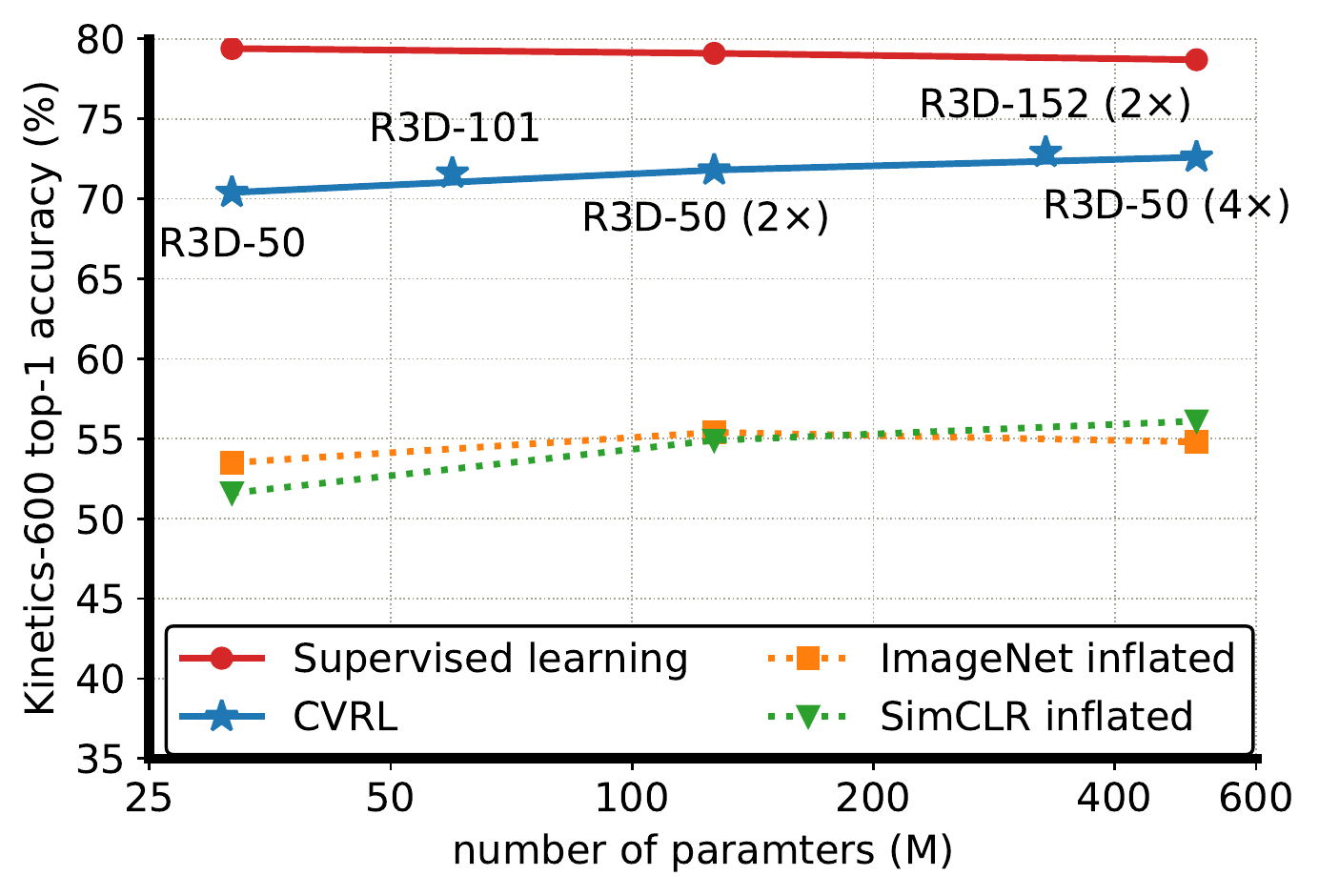}
\caption{\textbf{Kinetics-600 top-1 linear classification accuracy} of different spatiotemporal representations. CVRL outperforms ImageNet supervised~\cite{resnet} and SimCLR unsupervised~\cite{simclr} pre-training using the same 3D inflated ResNets, closing the gap between unsupervised and supervised video representation learning.}
\label{fig:teaser}
\vspace{-2mm}
\end{figure}

\section{Introduction}
Representation learning is of crucial importance in computer vision tasks, and a number of highly promising recent developments in this area have carried over successfully from the static image domain to the video domain.
Classic hand-crafted local invariant features (\eg, SIFT~\cite{sift}) for images have their counterparts (\eg, 3D SIFT~\cite{3dsift}) in videos, where the temporal dimension of videos gives rise to key differences between them. 
Similarly, state-of-the-art  neural networks for video understanding~\cite{c3d,i3d,resnet3d,s3d,slowfast,x3d} often extend 2D convolutional neural networks~\cite{resnet,mobilenets} for images along the temporal dimension. 
More recently, unsupervised or self-supervised learning of representations from unlabeled visual data~\cite{moco,simclr,byol,swav} has gained momentum in the literature partially thanks to its ability to model the abundantly available unlabeled data.

However, self-supervised learning gravitates to different dimensions in videos and images, respectively. It is natural to engineer self-supervised learning signals along the temporal dimension in videos. 
Examples abound, including models for predicting the future~\cite{srivastava2015unsupervised,lotter2016deep,han2019video}, changing temporal sampling rates~\cite{yang2020video}, sorting video frames or clips~\cite{lee2017unsupervised,kim2019self,xu2019self} and combining a few tasks~\cite{bai2020can}.
Meanwhile, in the domain of static images, some recent work~\cite{moco,simclr,byol,swav} that exploits spatial self-supervision has reported unprecedented performance on image representation learning. 

\begin{figure*}[t]
\centering
\includegraphics[width=0.9\textwidth]{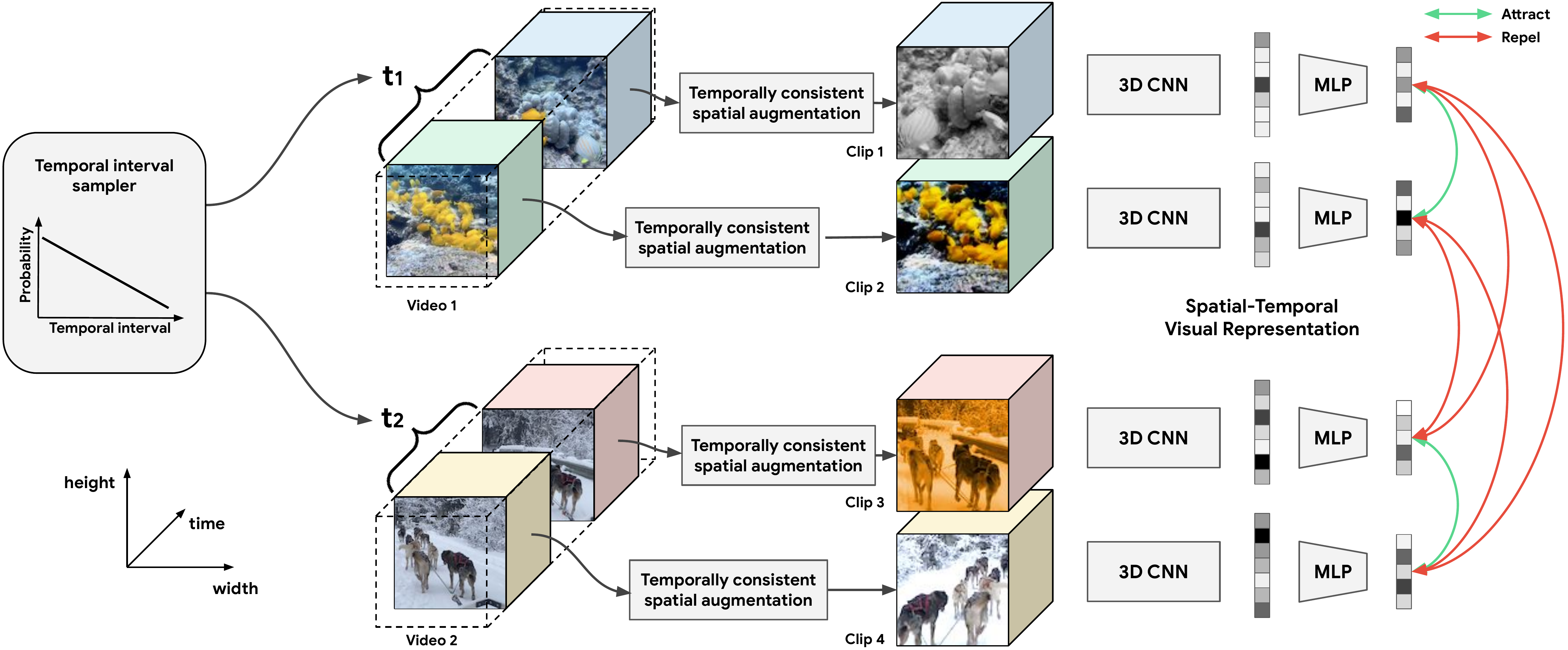}
\caption{\textbf{Overview of the proposed spatiotemporal Contrastive Video Representation Learning (CVRL) framework.} From a raw video, we first sample a temporal interval from a monotonically decreasing distribution. The temporal interval represents the number of frames between the start points of two clips, and we sample two clips from a video according to this interval. Afterwards we apply a temporally consistent spatial augmentation to each of the clips and feed them into a 3D backbone with an MLP head. The contrastive loss is used to train the network to attract the clips from the same video and repel the clips from different videos in the embedding space.}
\label{fig:overview}
\vspace{2mm}
\end{figure*}

The long-standing pursuit after temporal cues for self-supervised video representation learning has left self-supervision signals in the spatial subspace under-exploited for videos. 
To promote the spatial self-supervision signals in videos, we build a Contrastive Video Representation Learning (CVRL) framework to learn spatiotemporal representations from unlabeled videos. Figure~\ref{fig:overview} illustrates our framework, which contrasts the similarity between two positive video clips against those of negative pairs using the InfoNCE contrastive loss~\cite{cpc}. Since there is no label in self-supervised learning, we construct positive pairs as two augmented video clips sampled from the same input video. 

We carefully design data augmentations to involve both spatial and temporal cues for CVRL.
Simply applying spatial augmentation independently to video frames actually hurts the learning because it breaks the natural motion along the time dimension. 
Instead, we propose a  temporally consistent spatial augmentation method by fixing the randomness across frames. It is simple and yet vital as demonstrated in our experiments. 
For temporal augmentation, we take visual content into account by a sampling strategy tailored for the CVRL framework. 
On the one hand, a pair of positive clips that are temporally distant may contain very different visual content, leading to a low similarity that could be indistinguishable from those of the negative pairs. 
On the other hand, completely discarding the clips that are far in time reduces the temporal augmentation effect. To this end, we propose a sampling strategy to ensure the time difference between two positive clips follows a monotonically decreasing distribution. Effectively, CVRL mainly learns from positive pairs of temporally close clips and secondarily sees some temporally distant clips during training. The efficacy of the proposed spatial and temporal augmentation methods is verified by extensive ablation studies.

We primarily evaluate the learned video representations on both Kinetics-400~\cite{kay2017kinetics} and Kinetics-600~\cite{kinetics600} by training a linear classifier following~\cite{simclr,moco} on top of frozen backbones. 
We also study semi-supervised learning, downstream action classification and detection to further assess CVRL. 
We next summarize our main findings.

\vspace{-1mm}
\paragraph{Mixing spatial and temporal cues boosts the performance.} Relying on spatial or temporal augmentation only yields relatively low performance, as shown in Table~\ref{tab:ablation}. In contrast, we achieve an improvement of 22.9\% top-1 accuracy by combining both augmentations in the manner we proposed above, \ie, temporally consistent spatial augmentation and the temporal sampling strategy.

\vspace{-1mm}
\paragraph{Our representations outperform prior arts.} The linear evaluation of CVRL achieves more than 15\% gain over competing baselines, as shown in Figure~\ref{fig:teaser} and Table~\ref{tab:linear_eval}. 
On Kinetics-400, CVRL achieves 12.6\% improvement over ImageNet pre-training, which were shown competitive in previous work~\cite{yang2020video,gordon2020watching}. For semi-supervised learning (Table~\ref{tab:semi}), CVRL surpasses all other baselines especially when there is only 1\% labeled data, indicating the advantage of our self-learned feature is more profound with limited labels. For downtream action classification on UCF-101~\cite{ucf101} and HMDB-51~\cite{kuehne2011hmdb}, CVRL has obvious advantages over other methods based on the vision modality and is competitive with state-of-the-art multimodal methods (Table~\ref{tab:ucf}).

\vspace{-1mm}
\paragraph{Our CVRL framework benefits from larger datasets and networks.} We study the effect of more training data in CVRL. We design an evaluation protocol by first pre-training models on different amounts of data with same iterations, and then comparing the performance on the same validation set. As shown in Figure~\ref{fig:scala}, a clear improvement is observed by using 50\% more data, demonstrating the potential of CVRL to scale to larger unlabeled datasets. We also conduct experiments with wider \& deeper networks and observe consistent improvements (Table~\ref{tab:linear_eval}), demonstrating that CVRL is more effective with larger networks.\\

\section{Related Work} \label{sec:related}

\paragraph{Self-supervised video representation learning.}
It is natural to exploit the temporal dimension in self-supervised video representation learning. Some early work predicts the future on top of frame-wise representations~\cite{srivastava2015unsupervised}. More recent work learns from raw videos by predicting motion and appearance statistics~\cite{wang2019self}, speed~\cite{benaim2020speednet, wang2020self} and encodings~\cite{lotter2016deep,han2019video,han2020memory}.
Aside from predicting the future, other common approaches include sorting frames or video clips~\cite{lee2017unsupervised,xu2019self,kim2019self,fernando2017self} along the temporal dimension and learning from proxy tasks like rotation~\cite{jing2018self}. Yang~\etal~\cite{yang2020video} learn by maintaining consistent representations of  different sampling rates. 
Furthermore, videos can often supply multimodal signals for cross-modality self-supervision, such as geometric cues~\cite{gan2018geometry}, speech or language~\cite{sun2019videobert, sun2019learning, miech2020end}, audio~\cite{korbar2018cooperative, alwassel2019self, patrick2020multi, asano2020labelling}, optical flow~\cite{han2020coclr} or combinations of multiple modalities ~\cite{alayrac2020self} and tasks~\cite{Piergiovanni_2020_CVPR}.

\vspace{-1mm}
\paragraph{Self-supervised image representation learning.}
Some early work learns visual representations from unlabeled images via manually specified pretext tasks, for instance, the auto-encoding methods~\cite{pathak2016context,zhang2016colorful,zhang2017split} that leverage contexts, channels, or colors. 
Other pretext tasks include but are not limited to relative patch location~\cite{doersch2015unsupervised}, jigsaw puzzles~\cite{noroozi2016unsupervised}, and image rotations~\cite{gidaris2018unsupervised}.
Interestingly, most of the pretext tasks can be integrated into a contrastive learning framework~\cite{moco,simclr,ye2019unsupervised,byol,cpc,henaff2019data,tian2019contrastive}, which maintains relative consistency between the representations of an image and its augmented view. The augmentation could encompass various pretext tasks.
Tian~\etal~\cite{tian2020makes} study what makes a good view in this framework.
Clustering can also provide an effective addition to the framework~\cite{swav}. It is worth noting that the recent wave of contrastive learning shares a similar loss objective as instance discrimination~\cite{wu2018unsupervised}. 

\vspace{-1mm}
\paragraph{Videos as supervision for images and beyond.}
Video can help supervise the learning of image representations~\cite{wang2015unsupervised,pathak2017learning,vondrick2018tracking,gordon2020watching,Purushwalkam12020Demystifying}, correspondences~\cite{CVPR2019_CycleTime,dwibedi2019temporal}, and robotic behaviors~\cite{sermanet2018time} thanks to its rich content about different views of objects and its motion and tracking cues. 
On the other hand, Girdhar~\etal~\cite{girdhar2019distinit} propose to learn video representations by distillation from image representations.

\section{Methodology}
\subsection{Video Representation Learning Framework}
We build our self-supervised contrastive video representation learning framework as illustrated in Figure~\ref{fig:overview}. The core of this framework is an InfoNCE contrastive loss~\cite{cpc} applied on features extracted from augmented videos. 
Suppose we sample $N$ raw videos and augment them, resulting in $2N$ clips (the augmentation module is described in Section~\ref{sec:augmentation}).
Denote $\bm{z}_i, \bm{z}_i'$ as the encoded representations of the two augmented clips of the $i$-th input video. The InfoNCE contrastive loss is defined as $\mathcal{L} =\frac{1}{N}\sum_{i=1}^N \mathcal{L}_{i}$ and
\begin{equation}
    \mathcal{L}_i=-\log\frac{\exp(\text{sim}(\bm{z}_i,\bm{z}_i')/\tau)}{\sum_{k=1}^{2N}\mathbf{1}_{[k\neq i]}\exp(\text{sim}(\bm{z}_i,\bm{z}_k)/\tau)}, \label{eq:infonce}
\end{equation}
where $\text{sim}(\bm{u},\bm{v})=\bm{u}^\top\bm{v}/\|\bm{u}\|_2\|\bm{v}\|_2$ is the inner product between two $\ell_2$ normalized vectors, $\mathbf{1}_{[\cdot]}$ is an indicator excluding  from the denominator the self-similarity of the encoded video $\bm{z}_i$, and $\tau>0$ is a temperature parameter. The loss allows the positive pair $(\bm{z}_i, \bm{z}_i')$ to attract mutually while they repel the other items in the mini-batch. 

We discuss other components of the framework as follows: (1) an encoder network maps an input video clip to its representation $\bm{z}$, (2) spatiotemporal augmentations to construct positive pairs $(\bm{z}_i, \bm{z}_i')$ and the properties they induce, and (3) methods to evaluate the learned representations.

\subsection{Video Encoder}
\label{sec:encoder}
We encode a video sequence using 3D-ResNets~\cite{resnet, resnet3d} as backbones. 
We expand the original 2D convolution kernels to 3D to capture spatiotemporal information in videos. 
The design of our 3D-ResNets mainly follows the ``slow'' pathway of the SlowFast network~\cite{slowfast} with two minor modifications: (1) the temporal stride of 2 in the data layer, and (2) the temporal kernel size of 5 and stride of 2 in the first convolution layer. We also take as input a higher temporal resolution. Table~\ref{tab:network} and Section~\ref{sec:implementation-details} provide more details of the network.
The video representation is a 2048-dimensional feature vector. 
As suggested by SimCLR~\cite{simclr}, we add a multi-layer projection head onto the backbone to obtain the encoded 128-dimensional feature vector $\bm{z}$ used in Equation~\ref{eq:infonce}. 
During evaluation, we discard the MLP and use the 2048-dimensional representation directly from the backbone to make the video encoder compatible with other supervised learning methods. We also experiment with
$2\times$ and $4\times$ backbones, which multiply the number of filters in the network, including the backbone's output feature dimension and all layers in MLP, by $2\times$ and $4\times$ accordingly.

\newcommand{\blocks}[3]{\multirow{3}{*}{\(\left[\begin{array}{c}\text{1$\times$1$^\text{2}$, #2}\\[-.1em] \text{1$\times$3$^\text{2}$, #2}\\[-.1em] \text{1$\times$1$^\text{2}$, #1}\end{array}\right]\)$\times$#3}
}
\newcommand{\blockt}[3]{\multirow{3}{*}{\(\left[\begin{array}{c}\text{\underline{3$\times$1$^\text{2}$}, #2}\\[-.1em] \text{1$\times$3$^\text{2}$, #2}\\[-.1em] \text{1$\times$1$^\text{2}$, #1}\end{array}\right]\)$\times$#3}
}

\begin{table}[t]
    \centering
    \resizebox{0.93\columnwidth}{!}{
        \tablestyle{5pt}{1.1}
        \begin{tabular}{c|c|c}
            \shline
            Stage & Network & Output size $T \times S^2$ \\
            \shline
            raw clip & - & $32\times224^\text{2}$ \\
            \hline
            data & stride \textbf{2}, 1$^\text{2}$ & $16\times224^\text{2}$ \\
            \hline
            \multirow{2}{*}{conv$_1$} & \multicolumn{1}{c|}{\underline{$\textbf{5}\times7^\text{2}$}, {64}} & \multirow{2}{*}{$8\times112^\text{2}$}  \\
            & stride \textbf{2}, 2$^\text{2}$ & \\
            \hline
            \multirow{2}{*}{pool$_1$}  & \multicolumn{1}{c|}{$1\times3^\text{2}$ max} &  \multirow{2}{*}{$8\times56^\text{2}$} \\
            & stride 1, 2$^\text{2}$ & \\
            \hline
            \multirow{3}{*}{conv$_2$} & \blocks{{256}}{{64}}{3} & \multirow{3}{*}{$8\times56^\text{2}$}  \\
            &  & \\
            &  & \\
            \hline
            \multirow{3}{*}{conv$_3$} & \blocks{{512}}{{128}}{4} & \multirow{3}{*}{$8\times28^\text{2}$}  \\
            &  & \\
            &  & \\
            \hline
            \multirow{3}{*}{conv$_4$} & \blockt{{1024}}{{256}}{6} & \multirow{3}{*}{$8\times14^\text{2}$} \\
            &  & \\
            &  & \\
            \hline
            \multirow{3}{*}{conv$_5$} & \blockt{{2048}}{{512}}{3} & \multirow{3}{*}{$8\times7^\text{2}$} \\
            &  & \\
            &  & \\
            \hline
			\multicolumn{2}{c|}{global average pooling} &  $1\times1^\text{2}$ \\
			\shline
    \end{tabular}}
    \vspace{2mm}
    \caption{\textbf{Our video encoder: a 3D-ResNet-50 (R3D-50).} The input video has 16 frames (stride 2) in self-supervised pre-training and 32 frames (stride 2) in linear evaluation, semi-supervised learning, supervised learning and downstream tasks.}
    \label{tab:network}
\vspace{-2mm}
\end{table}

\subsection{Data Augmentation}
\label{sec:augmentation}
\begin{figure}[t]
\centering
\subfigure[$P(t) \propto -t + c$ (63.8\% acc.)]{
\includegraphics[width=0.225\textwidth]{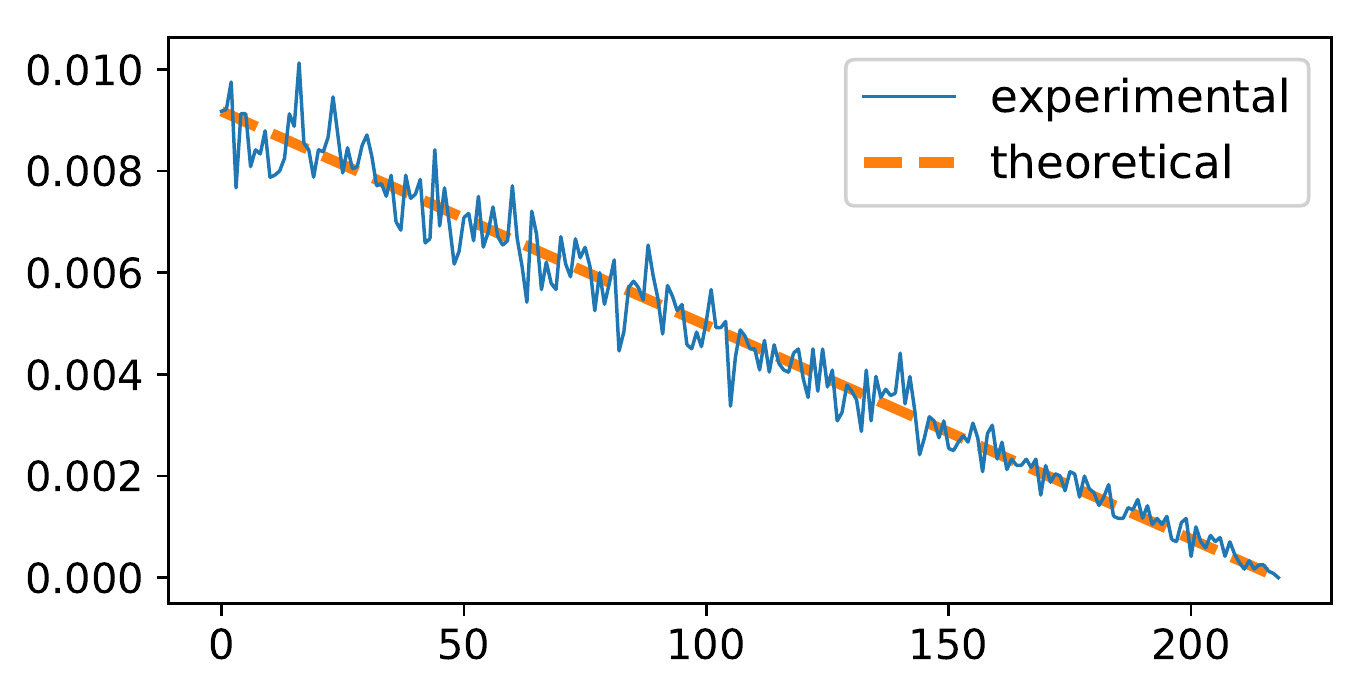}
\label{dis:1}
} \hfill
\subfigure[$P(t) \propto -t^{0.5} + c$ (63.1\% acc.)]{
\includegraphics[width=0.225\textwidth]{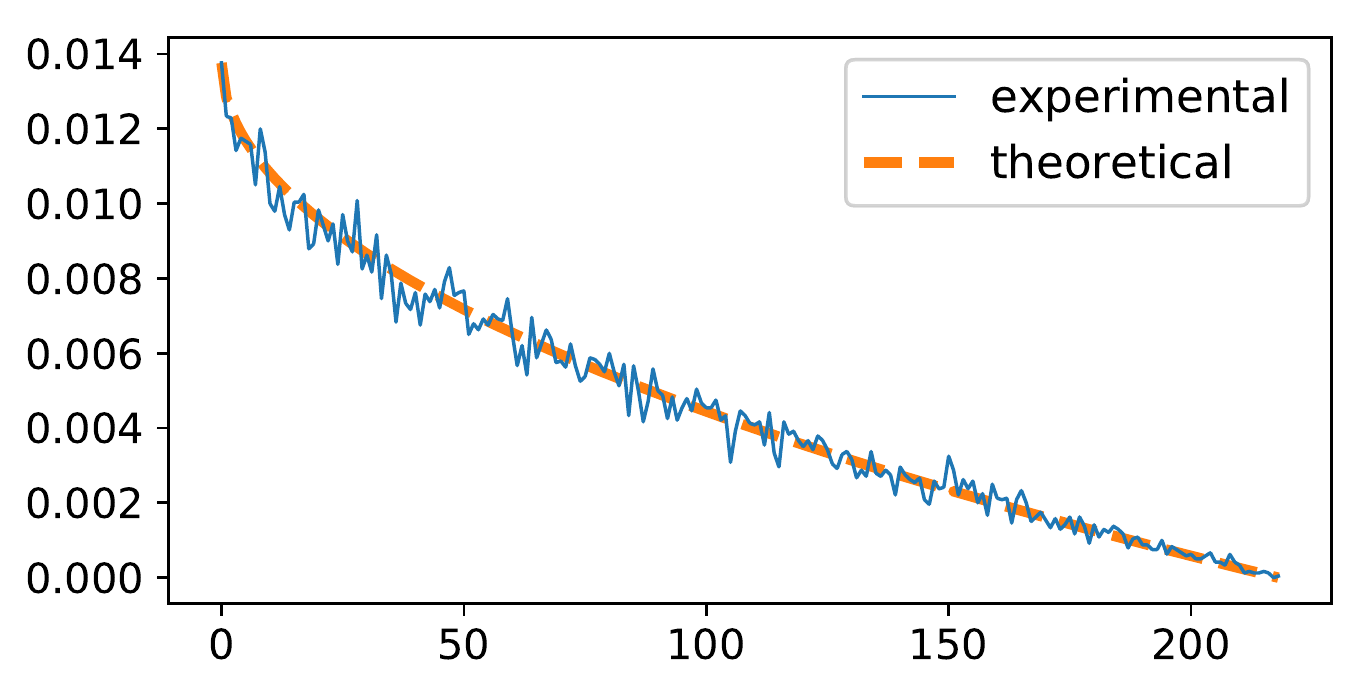}
\label{dis:2}
}\vspace{-3mm}
\subfigure[$P(t) \propto -t^2 + c$ (62.9\% acc.)]{
\includegraphics[width=0.225\textwidth]{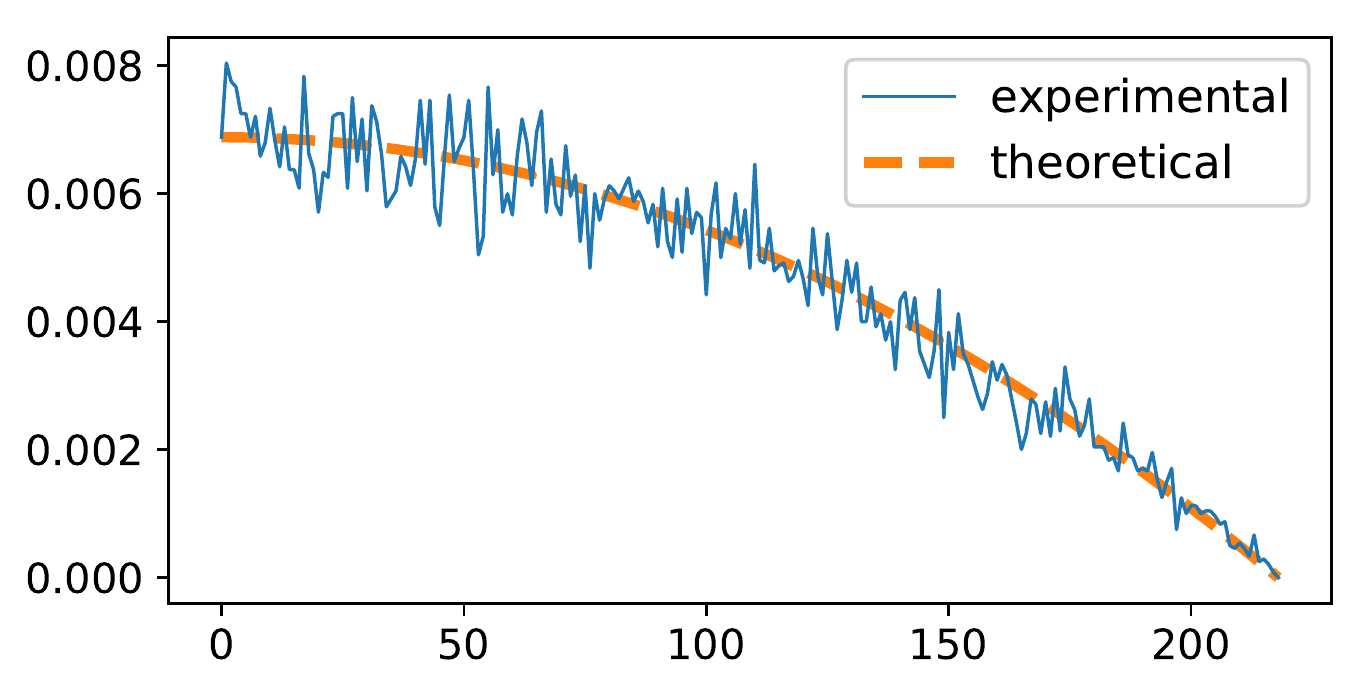}
\label{dis:3}
} \hfill
\subfigure[$P(t) \propto c$ (62.7\% acc.)]{
\includegraphics[width=0.225\textwidth]{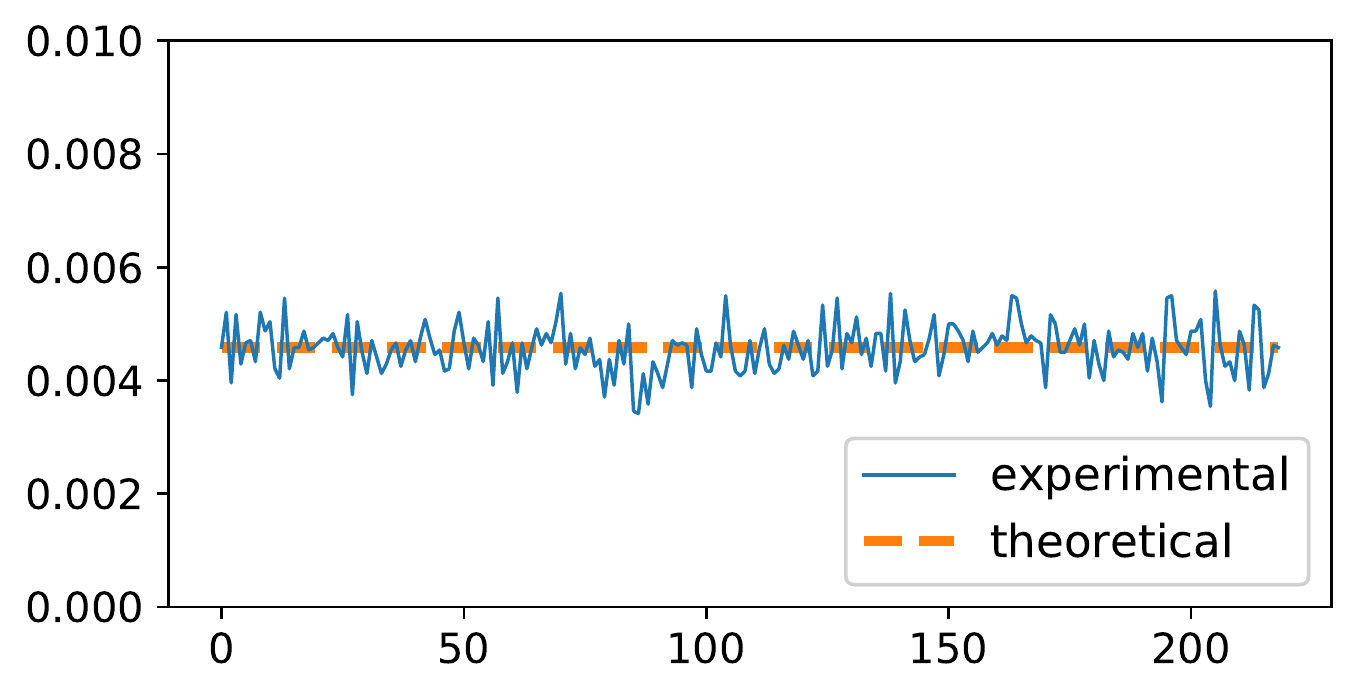}
\label{dis:4}
}\vspace{-3mm}
\subfigure[$P(t) \propto t + c$ (62.4\% acc.)]{
\includegraphics[width=0.225\textwidth]{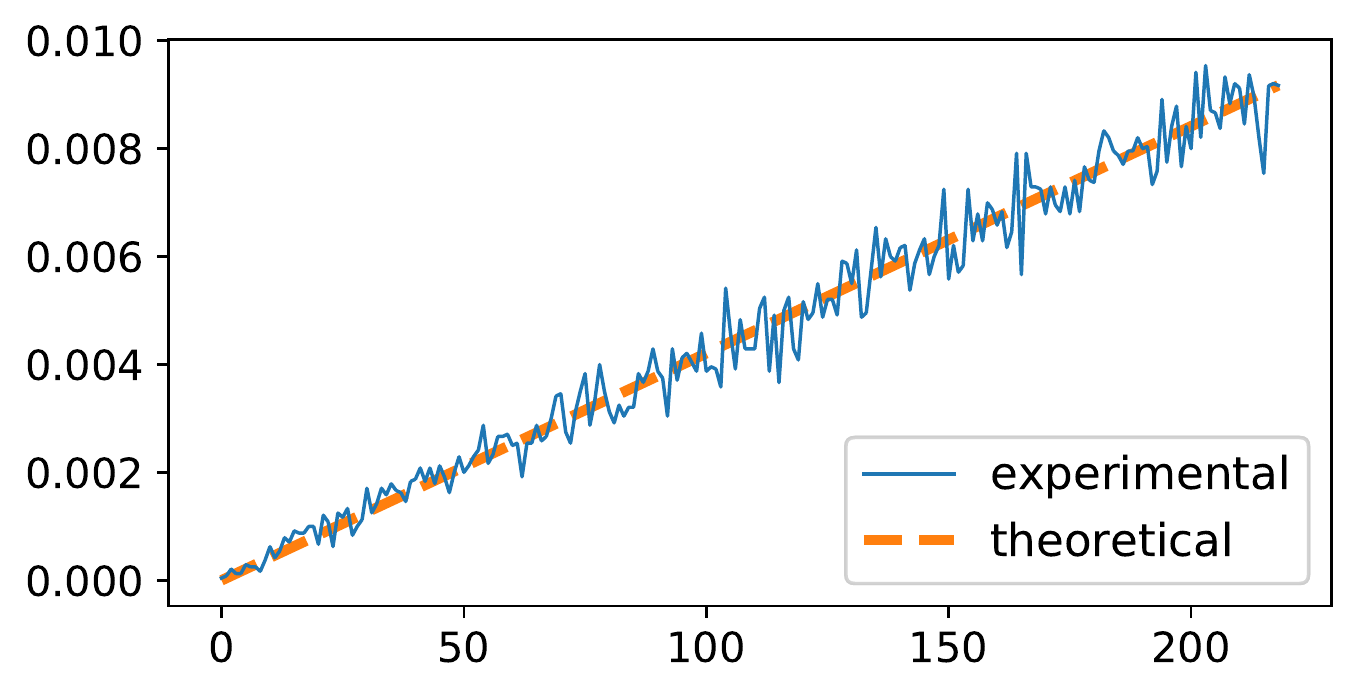}
\label{dis:5}
} \hfill
\subfigure[$P(t) \propto t^2 + c$ (61.9\% acc.)]{
\includegraphics[width=0.225\textwidth]{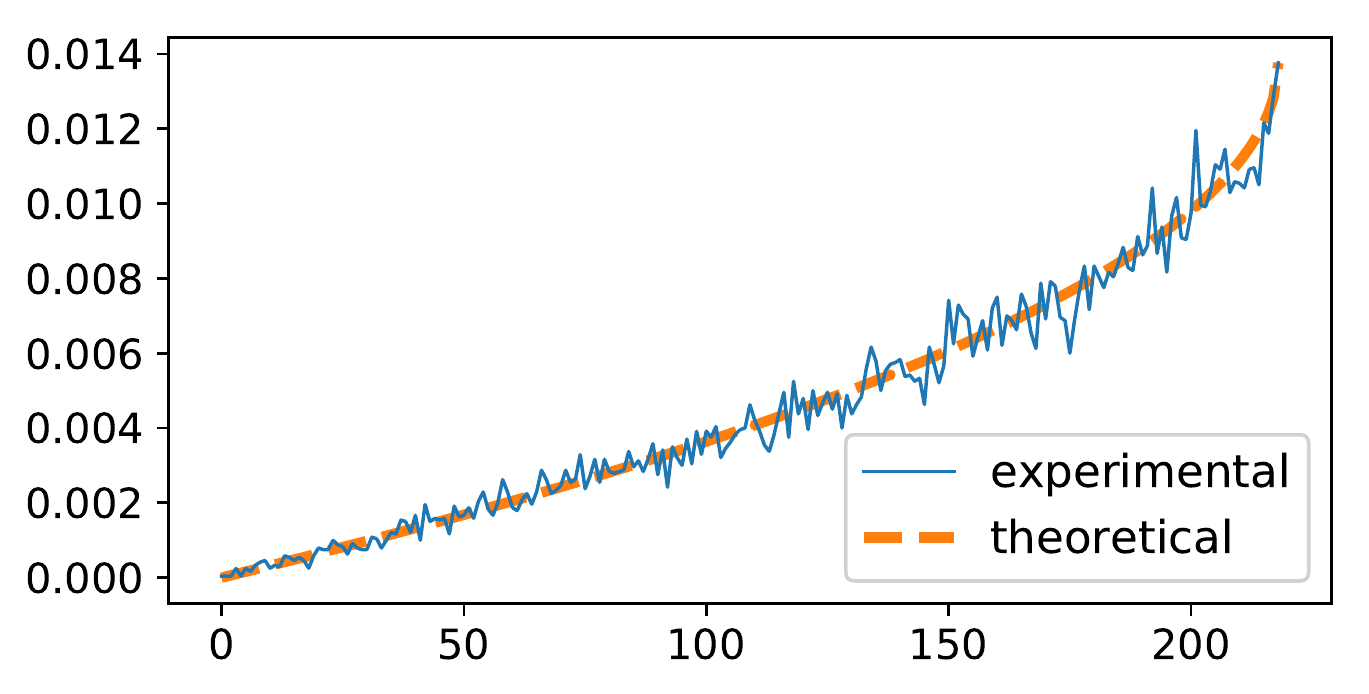}
\label{dis:6}
}
\caption{\textbf{Performance of different sampling distributions.} The x-axis is the temporal interval $t$ between two clips in a video, and the y-axis is the sampling probability $P(t)$. We report linear evaluation accuracy upon 200 epochs of pre-training on Kinetics-400.}
\label{fig:sampling_dis}
\vspace{-2mm}
\end{figure}

The flexibility of CVRL allows us to study a variety of desired properties, which are incorporated in the form of data augmentations.
We focus on the augmentations in both temporal and spatial dimensions.
\vspace{-1mm}
\paragraph{Temporal Augmentation: a sampling perspective.}
It is straightforward to take two clips from an input video as a positive pair, but how to sample the two clips matters.
Previous work provides temporal augmentation techniques like sorting video frames or clips~\cite{lee2017unsupervised,kim2019self,xu2019self}, altering playback rates~\cite{Yao_2020_CVPR, wang2020self},~\etc.  However, directly incorporating them into CVRL would result in learning temporally invariant features, which opposes the temporally evolving nature of videos. 
We instead account for the temporal changes using a sampling strategy. The main motivation is that two clips from the same video would be more distinct when their temporal interval is larger. If we sample temporally distant clips with smaller probabilities, the contrastive loss (Equation~\ref{eq:infonce}) would focus more on the temporally close clips, pulling their features closer and imposing less penalty over the clips that are far away in time.
Given an input video of length $T$, our sampling strategy takes two steps. We first draw a time interval $t$ from a distribution $P(t)$ over $[0,T]$. We then uniformly sample a clip from $[0,T-t]$, followed by the second clip which is delayed by $t$ after the first. 
More details on the sampling procedure can be found in Appendix~\ref{appdix:sampling}.
We experiment with monotonically increasing, decreasing, and uniform distributions, as illustrated in Figure~\ref{fig:sampling_dis}. We find that decreasing distributions~(a-c) generally perform better than the uniform~(d) or increasing ones~(e-f), aligning well with our motivation above of assigning lower sampling probability on larger temporal intervals.

\vspace{-1mm}
\paragraph{Spatial Augmentation: a temporally consistent design.}
Spatial augmentation is widely used in both supervised learning and unsupervised learning for images. 
Although the question of how to apply strong spatial augmentations to videos remains open, a natural strategy is to utilize existing image-based spatial augmentation methods to the video frames one by one.
However, this method could break the motion cues across frames.
Spatial augmentation methods often contain some randomness such as random cropping, color jittering and blurring as important ways to strengthen their effectiveness.
In videos, however, such randomness between consecutive frames, could negatively affect the representation learning along the temporal dimension.
Therefore, we design a simple yet effective approach to address this issue, by making the spatial augmentations consistent along the temporal dimension.
With fixed randomness across frames, the 3D video encoder is able to better utilize spatiotemporal cues. 
This approach is validated by experimental results in Table~\ref{tab:ablation}.
Algorithm~\ref{alg:aug} demonstrates the detailed procedure of our temporally consistent spatial augmentations, where the hyper-parameters are only generated once for each video and applied to all frames. An illustration can be found in Appendix~\ref{appdix:tmp_consis}.

\subsection{Evaluation}
As a common practice in self-supervised representation learning~\cite{simclr,moco}, we mainly evaluate the learned video representations by fixing the weights in the pre-trained video encoder and training a linear classifier on top of it. We also assess the learned representations by fine-tuning the entire video encoder network in a semi-supervised learning setting as well as in downstream action classification and detection tasks. More details to come in Section~\ref{sec:exp}.

\begin{algorithm}[t]
\small
\caption{\label{alg:aug} \small{Temporally consistent spatial augmentation}}
\begin{algorithmic}
\REQUIRE Video clip $V = \{f_1, f_2, \cdots, f_M\}$ with $M$ frames
\STATE \textbf{Crop:} Randomly crop a spatial region with size ratio \textbf{S} \\ 
in range of $[0.3, 1]$ and aspect ratio \textbf{A} in $[0.5, 2]$
\STATE \textbf{Resize:} Resize the cropped region to size of $224 \times 224$
\STATE \textbf{Flip:} Draw a flag $\textbf{F}_f$ from $\{0, 1\}$ with 50\% on $1$
\STATE \textbf{Jitter:} Draw a flag $\textbf{F}_j$ from $\{0, 1\}$ with 80\% on $1$
\STATE \textbf{Grey:} Draw a flag $\textbf{F}_g$ from $\{0, 1\}$ with 20\% on $1$
\FOR{$k\in \{1, \ldots, M\}$}
\STATE $~~~~$ $f'_k = \text{Resize}\big(\text{Crop}(f_k, size=\textbf{S}, aspect=\textbf{A})\big)$
\STATE $~~~~$ $f'_k = \text{Flip}(f'_k)$ if $\textbf{F}_f = 1$
\STATE $~~~~$ $f'_k = \text{Color\_jitter}(f'_k)$ if $\textbf{F}_j = 1$
\STATE $~~~~$ $f'_k = \text{Greyscale}(f'_k)$ if $\textbf{F}_g = 1$
\STATE $~~~~$ $f'_k = \text{Gaussian\_blur}(f'_k)$ 
\ENDFOR
\ENSURE Augmented video clip $V' = \{f'_1, f'_2, \cdots, f'_M\}$
\end{algorithmic}
\end{algorithm}
\vspace{-2mm}

\begin{table*}[t]
\small
\centering
\begin{tabular}{lrrccc}
\shline
Method & Backbone (\#params) & Pre-train data (duration) & Mod. & Linear eval. & Top-1 Acc. (\%) \\
\shline
VTHCL~\cite{yang2020video} & R3D-50 (31.7M) & K400 (28d) & V & K400 & 37.8 \\
SimCLR inflated & R3D-50 (31.7M) & K400 (28d) & V & K400 & 46.8 \\ 
VINCE~\cite{gordon2020watching} & R-50 (23.5M) & K400 (28d) & V & K400 & 49.1 \\
ImageNet inflated & R3D-50 (31.7M) & ImageNet (N/A) & V & K400 & 53.5\\
SeCo~\cite{yao2020seco} & R-50 (23.5M) & K400 (28d) & V & K400 & 61.9 \\
\hline
CVRL & R3D-50 (31.7M) & K400 (28d) & V & K400 & \textbf{66.1}\\
CVRL & R3D-101 (59.7M) & K400 (28d) & V & K400 & \textbf{67.6}\\
CVRL & R3D-152 (2$\times$) (328.0M) & K600 (44d) & V & K400 & \textbf{71.6}\\
\hline
\textcolor{gray}{Supervised (K400)} & \textcolor{gray}{R3D-50 (31.7M)} & \textcolor{gray}{N/A} & \textcolor{gray}{V} & \textcolor{gray}{N/A} & \textcolor{gray}{76.0}\\
\shline
SimCLR inflated & R3D-50 (31.7M) & K600 (44d) & V & K600 & 51.6\\ 
ImageNet inflated & R3D-50 (31.7M) & ImageNet (N/A) & V & K600 & 54.7\\
MMV-VA~\cite{alayrac2020self} & S3D-G (9.1M) & AS + HT (16y) & VA & K600 & 59.8\\
MMV~\cite{alayrac2020self} & TSM-50$\times$2 (93.9M) & AS + HT (16y) & VAT & K600 & 70.5\\
\hline
CVRL & R3D-50 (31.7M) & K600 (44d) & V & K600 & \textbf{70.4}\\
CVRL & R3D-101 (59.7M) & K600 (44d) & V & K600 & \textbf{71.6}\\
CVRL & R3D-152 (2$\times$) (328.0M) & K600 (44d) & V & K600 & \textbf{72.9}\\
\hline
\textcolor{gray}{Supervised (K600)} & \textcolor{gray}{R3D-50 (31.7M)} & \textcolor{gray}{N/A} & \textcolor{gray}{V} & \textcolor{gray}{N/A} & \textcolor{gray}{79.4}\\
\shline
\end{tabular}
\vspace{2mm}
\caption{\textbf{Linear evaluation results.} CVRL shows superior performance compared to state-of-the-art methods and baselines, significantly closes the gap with supervised learning. R-50 in the network column represents the standard 2D ResNet-50.}
\label{tab:linear_eval}
\end{table*}

\section{Experiments}
\label{sec:exp}
We mainly conduct experiments on the Kinetics-400 (K400)~\cite{kay2017kinetics} and Kinetics-600~\cite{kinetics600} (K600) datasets. 
K400 consists of about 240k training videos and 20k validation videos belonging to 400 action classes. 
K600 is a superset of K400 by revising ambiguous classes and adding 200 more classes, containing about 360k training and 28k validation videos from 600 classes.
We note that K400 has been extensively used in the literature and hope our additional results on K600 would further demonstrate the effectiveness of CVRL and offer a reference to the field.
The videos in Kinetics have a duration of around 10 seconds, with 25 frames per second (\ie, around 250 frames per video). We adopt the standard protocol~\cite{simclr, moco} of self-supervised pre-training and linear evaluation as the primary metric for evaluating the learned representations. We also evaluate the learned representations via semi-supervised learning and downstream tasks. 

\subsection{Implementation Details}
\label{sec:implementation-details}
We use SGD as our optimizer with the momentum of 0.9. All models are trained with the mini-batch size of 1024 except for downstream tasks. We linearly warm-up the learning rate in the first 5 epochs~\cite{goyal2017accurate} followed by the scheduling strategy of half-period cosine learning rate decay~\cite{he2019bag}.
We apply the proposed temporal and spatial augmentations for the self-supervised pre-training.
For other tasks, we only use standard data augmentations of cropping, resizing, and flipping.
During testing, we densely sample 10 clips from each video and apply a 3-crop evaluation following~\cite{slowfast}.

\vspace{-1mm}
\paragraph{Self-supervised pre-training.}
We sample two 16-frame clips with the temporal stride of 2 from each video for the self-supervised pre-training of video representations. The duration of a clip is 1.28 seconds out of around 10 seconds of a video. We use synchronized batch normalization to avoid information leakage or overfitting~\cite{simclr}. The temperature $\tau$ is set to 0.1 in the InfoNCE loss for all experiments. The initial learning rate is set to 0.32.

\vspace{-1mm}
\paragraph{Linear evaluation.}
We evaluate video representations using a linear classifier by fixing all the weights in the backbone. During training, we sample a 32-frame clip with the temporal stride of 2 from each video to train the linear classifier for 100 epochs with an initial learning rate of 32. We $\ell_2$ normalize the feature before feeding it to the classifier. 

\vspace{-1mm}
\paragraph{Semi-supervised learning.}
We conduct semi-supervised learning, namely, by fine-tuning the pre-trained network on small subsets of Kinetics.
We sample 1\% and 10\% videos from each class in the training set, forming two balanced subsets, respectively. The evaluation set remains the same.
We use pre-trained backbones to initialize network parameters and fine-tune all layers using an initial learning rate of 0.2 without warm-up.
We train the model for 100 epochs on the 1\% subset and 50 epochs on the 10\% subset.

\vspace{-1mm}
\paragraph{Downstream action classification.} 
On UCF-101~\cite{ucf101} and HMDB-51~\cite{kuehne2011hmdb}, we use the pre-trained backbone on Kinetics to initialize the network parameters.
We report results for both fine-tuning (\ie, fine-tune all layers) and linear evaluation (\ie, train a linear classifier by fixing all backcbone weights). 
All the models are trained with a mini-batch size of 128 for 50 epochs.
We use an initial learning rate of 0.16 for fine-tuning on both datasets, 0.8 for linear evaluation on UCF-101 and 0.2 for HMDB-51. 

\vspace{-1mm}
\paragraph{Downstream action detection.} 
For action detection, we work on AVA~\cite{gu2018ava} containing 211k training and 57k validation videos. AVA provides spatiotemporal labels of each action in long videos of 15 to 30 minutes. Following~\cite{slowfast, yang2020video}, we adopt a Faster-RCNN~\cite{ren2015faster} baseline with modifications to enable it to process videos. We use pre-trained backbones on Kinetics-400 to initialize the detector, and train with the standard 1$\times$ schedule (12 epochs, decay learning rate by 10$\times$ at 8-th and 11-th epoch). We use an initial learning rate of 0.2 with 32 videos per batch.

\vspace{-1mm}
\paragraph{Supervised learning.}
To understand where CVRL stands, we also report supervised learning results. The setting for supervised learning is the same as linear evaluation except that we train the entire encoder network from scratch for 200 epochs without feature normalization. We use an initial learning rate of 0.8 and a dropout rate of 0.5 following~\cite{slowfast}.

\subsection{Experimental Results}

\paragraph{Comparison baselines.}
We compare our CVRL method with two baselines: (1) \textbf{ImageNet inflated}~\cite{resnet}: inflating the 2D ResNets pre-trained on ImageNet to our 3D ResNets by duplicating it along the temporal dimension, and (2) \textbf{SimCLR inflated}~\cite{simclr}: inflating the 2D ResNets pre-trained with SimCLR on the frame images of Kinetics~\footnote{We find a SimCLR model pre-trained on Kinetics frames slightly outperforms the same model pre-trained on ImageNet released by~\cite{simclr}. This is probably due to the domain difference between ImageNet and Kinetics.}. SimCLR inflated serves as an important frame-based baseline for our method by directly applying the state-of-the-art image self-supervised learning algorithm to Kinetics frames, where no temporal information is learned. In addition, we present the results of supervised learning as an upper bound of our method.

\paragraph{Notations.} We aim at providing an extensive comparison with prior work, but video self-supervised learning methods could be diverse in pre-training datasets and input modalities. For pre-training datasets, we use K400 in short for Kinetics-400~\cite{kay2017kinetics}, K600 for Kinetics-600~\cite{kinetics600}, HT for HowTo100M~\cite{miech2019howto100m}, AS for AudioSet~\cite{gemmeke2017audio}, IG65M for Instagram65M~\cite{ghadiyaram2019large}, and YT8M for YouTube8M~\cite{abu2016youtube}. We also calculate the total length of the videos in one dataset to indicate its scale, namely duration in the table, by using \textbf{y}ears or \textbf{d}ays. Following~\cite{alayrac2020self}, we divide modalities into four types: \textbf{V}ision, \textbf{F}low, \textbf{A}udio and \textbf{T}ext.

\paragraph{Linear evaluation.} Linear evaluation is the most straightforward way to quantify the quality of the learned representation. As shown in Table~\ref{tab:linear_eval}, while some previous state-of-the-art methods~\cite{yang2020video, gordon2020watching} are worse than ImageNet inflated, our CVRL outperforms the ImageNet inflated by 12.6\% in top-1 accuracy on K400. Compared with the frame-based SimCLR inflated encoder, CVRL has 19.7\% improvement, demonstrating the advantage of the learned spatiotemporal representation over spatial only ones. Finally, compared with the supervised upper bound, CVRL greatly closes the gap between self-supervised and supervised learning. We also compare CVRL with the very recent state-of-the-art multimodal video self-supervised learning method MMV~\cite{alayrac2020self} on K600. CVRL achieves performance that is on par with MMV~(70.4\% \vs 70.5\%), with 133$\times$ less pre-training data~(44 days \vs 16 years), 3$\times$ fewer parameters~(31.7M \vs 93.9M) and only a single vision modality~(V \vs VAT). With a deeper R3D-101, CVRL is able to show better performance~(71.6\% \vs 70.5\%) with only 60\% parameters~(59.7M \vs 93.9M). 
Pre-training and linear evaluation curves can be found in Appendix~\ref{appdix:stats}.

\definecolor{upcolor}{RGB}{0,0,0}
\begin{table}[t]
\small
\centering
\begin{tabular}{c|c|ll}
\shline
\multirow{2}{*}{Method} & \multicolumn{1}{c|}{\multirow{2}{*}{Backbone}} & \multicolumn{2}{c}{K400 Top-1 Acc. ($\Delta$ \vs Sup.)} \\
 \cline{3-4}
 & & \multicolumn{1}{c}{1\% label} & \multicolumn{1}{c}{10\% label} \\
\shline
{\hspace{-3mm}\begin{tabular}{c}Supervised\end{tabular}\hspace{-3mm}} & R3D-50 & ~~3.2 & 39.6 \\
{\hspace{-3mm}\begin{tabular}{c}SimCLR infla.\end{tabular}\hspace{-3mm}} & R3D-50 & 11.8 (\textcolor{upcolor}{8.6$\uparrow$}) & 46.1 (\textcolor{upcolor}{6.5$\uparrow$}) \\
{\hspace{-3mm}\begin{tabular}{c}ImageNet infla.\end{tabular}\hspace{-3mm}} & R3D-50 & 16.0 (\textcolor{upcolor}{12.8$\uparrow$}) & 49.1 (\textcolor{upcolor}{9.5$\uparrow$}) \\
\hline
{CVRL} & R3D-50 & \textbf{35.1} (\textcolor{upcolor}{\textbf{31.9}$\uparrow$})& \textbf{58.1} (\textcolor{upcolor}{\textbf{18.5}$\uparrow$}) \\
\shline
\end{tabular}
\vspace{2mm}
\caption{\textbf{Semi-supervised learning on Kinetics-400.}}
\label{tab:semi}
\vspace{-4mm}
\end{table}

\vspace{-1mm}
\paragraph{Semi-supervised learning.} For semi-supervised learning on K400, as presented in Table~\ref{tab:semi}, CVRL surpasses all other baselines across different architectures and label fractions, especially when there is only 1\% labeled data for fine-tuning, indicating that the advantage of our self-supervised CVRL is more profound when the labeled data is limited.
Results on K600 can be found in Appendix~\ref{appdix:semi_k600}.

\vspace{-1mm}
\paragraph{Downstream action classification.} Pre-training the network encoder on a large dataset and fine-tuning all layers or conducting linear evaluation on UCF-101~\cite{ucf101} and HMDB-51~\cite{kuehne2011hmdb} is the most common evaluation protocol in the video self-supervised learning literature. We organize previous methods mainly by (1) what input modality is used and (2) which dataset is pre-trained on. We provide a comprehensive comparison in Table~\ref{tab:ucf}. We first divide all entries by the input modality they used. Inside each modality, we arrange the entries \wrt the performance on UCF-101 by ascending order. We notice there is inconsistency in previous work on reporting results with different splits of UCF-101 and HMDB-51, so we report on both split-1 and 3 splits average.  For fine-tuning, CVRL significantly outperforms methods using \textbf{V}ision modality only. Compared with multimodal methods using \textbf{V}ision and \textbf{F}low~\cite{han2020memory, han2020coclr}, \textbf{V}ision and \textbf{T}ext ~\cite{miech2020end}, CVRL still ranks top. Multimodal methods using \textbf{V}ision and \textbf{A}udio are able to achieve better performance starting from GDT~\cite{patrick2020multi} on AS, while it is worth to point out that the pre-trained dataset is 9$\times$ larger than K400 which we pre-train CVRL on with single \textbf{V}ision modality. 
For CVRL pre-trained on K600, it is only worse than the best model of~\cite{alwassel2019self, patrick2020multi, Piergiovanni_2020_CVPR, alayrac2020self} on UCF-101 and outperforms the best model of~\cite{alwassel2019self, Piergiovanni_2020_CVPR} on HMDB-51, where their pre-training datasets are $108\times$ to $174\times$ larger than K600. For linear evaluation, CVRL is better than all single and multi modal methods with the only exception of MMV~\cite{alayrac2020self} on UCF-101. On HMDB-51, CVRL demonstrates very competitive performance, outperforming all methods using vision modality and multimodal methods of~\cite{han2020memory, han2020coclr, miech2020end, alwassel2019self}. In conclusion, CVRL shows competitive performance on downstream action classification, compared with single and mutilmodal video self-supervised learning methods.

\vspace{-1mm}
\paragraph{Downstream action detection.} We conduct experiments on AVA~\cite{gu2018ava} dataset which benchmarks methods for detecting when an action happens in the temporal domain and where it happens in the spatial domain. Each video in AVA is annotated for 15 to 30 minutes and we consider this as an important experiment to demonstrate the transferability of CVRL learned features. We adopt Faster-RCNN~\cite{ren2015faster} and replace the 2D ResNet backbone with our video encoder in Table~\ref{tab:network}. Following~\cite{slowfast,yang2020video}, we compute region-of-interest (RoI) features by using a 3D RoIAlign on the features from the last conv block. We then perform a temporal average pooling followed by a spatial max pooling, and feed the feature into a sigmoid-based classifier for mutli-label prediction. We use pre-trained weights to initialize the video encoder, and fine-tune all layers for 12 epochs. We report mean Average-Precision (mAP) in Table~\ref{tab:ava}, where CVRL shows better performance than baselines.

\begin{table}[t]
\footnotesize
\centering
\begin{tabular}{l|r|c|cc}
\shline
\multicolumn{1}{c|}{\multirow{2}{*}{Method}} & \multicolumn{1}{c|}{Pre-train data} & \multirow{2}{*}{\footnotesize{Mod.}} &\multicolumn{2}{c}{\footnotesize{Top-1 Acc. (\%)}} \\
& \multicolumn{1}{c|}{(duration)} &  & \footnotesize{UCF} & \footnotesize{HMDB} \\
\shline
\multicolumn{5}{c}{\multirow{2}{*}{\small{\textbf{Fine-Tuning}}}} \\
\multicolumn{5}{c}{\multirow{2}{*}{}} \\
\hline
MotionPred$^\dagger$~\cite{wang2019self} & K400 (28d) & V & 61.2 & 33.4 \\
3D-RotNet$^\ddagger$~\cite{jing2018self} & K400 (28d) & V & 64.5 & 34.3 \\
ST-Puzzle$^\ddagger$~\cite{kim2019self} & K400 (28d) & V & 65.8 & 33.7 \\
ClipOrder$^\dagger$~\cite{xu2019self} & K400 (28d) & V & 72.4 & 30.9 \\
DPC$^\ddagger$~\cite{han2019video} & K400 (28d) & V & 75.7 & 35.7 \\
PacePred$^\dagger$~\cite{wang2020self} & K400 (28d) & V & 77.1 & 36.6 \\
MemDPC$^\mathsection$~\cite{han2020memory} & K400 (28d) & V & 78.1 & 41.2 \\
CBT$^\ddagger$~\cite{sun2019learning} & K600+ (273d) & V & 79.5 & 44.6 \\
SpeedNet$^\dagger$~\cite{benaim2020speednet} & K400 (28d) & V & 81.1 & 48.8\\
VTHCL$^\mathsection$~\cite{yang2020video} & K400 (28d) & V & 82.1 & 49.2\\
DynamoNet$^\ddagger$~\cite{diba2019dynamonet} & YT8M(13y) & V & 88.1 & 59.9 \\
SeCo$^\mathsection$~\cite{yao2020seco} & K400 (28d) & V & 88.3 & 55.6\\
\hline
\textcolor{gray}{MemDPC$^\mathsection$~\cite{han2020memory}} & \textcolor{gray}{K400 (28d)}  & \textcolor{gray}{VF} & \textcolor{gray}{86.1} & \textcolor{gray}{54.5} \\
\textcolor{gray}{CoCLR$^\dagger$~\cite{han2020coclr}} & \textcolor{gray}{K400 (28d)}  & \textcolor{gray}{VF} & \textcolor{gray}{90.6} & \textcolor{gray}{62.9} \\
\hline
\textcolor{gray}{MIL-NCE$^\ddagger$~\cite{miech2020end}} & \textcolor{gray}{HT (15y)}  & \textcolor{gray}{VT} & \textcolor{gray}{91.3} & \textcolor{gray}{61.0} \\
\hline
\textcolor{gray}{AVTS$^\ddagger$~\cite{korbar2018cooperative}} & \textcolor{gray}{AS (240d)}  & \textcolor{gray}{VA} & \textcolor{gray}{89.0} & \textcolor{gray}{61.6} \\
\textcolor{gray}{MMV-VA$^\ddagger$~\cite{alayrac2020self}} & \textcolor{gray}{AS+HT (16y)}  & \textcolor{gray}{VA} & \textcolor{gray}{91.1} & \textcolor{gray}{68.3} \\
\textcolor{gray}{XDC$^\ddagger$~\cite{alwassel2019self}} & \textcolor{gray}{AS (240d)}  & \textcolor{gray}{VA} & \textcolor{gray}{91.2} & \textcolor{gray}{61.0} \\
\textcolor{gray}{GDT$^\ddagger$~\cite{patrick2020multi}} & \textcolor{gray}{AS (240d)}  & \textcolor{gray}{VA} & \textcolor{gray}{92.5} & \textcolor{gray}{66.1} \\
\textcolor{gray}{XDC$^\ddagger$~\cite{alwassel2019self}} & \textcolor{gray}{IG65M (21y)}  &
\textcolor{gray}{VA} & \textcolor{gray}{94.2} & \textcolor{gray}{67.4} \\
\textcolor{gray}{GDT$^\ddagger$~\cite{patrick2020multi}} & \textcolor{gray}{IG65M (21y)}  & \textcolor{gray}{VA} & \textcolor{gray}{95.2} & \textcolor{gray}{72.8} \\
\hline
\textcolor{gray}{Elo$^\mathsection$~\cite{Piergiovanni_2020_CVPR}} & \textcolor{gray}{YT8M (13y)}  & \textcolor{gray}{VFA} & \textcolor{gray}{93.8} & \textcolor{gray}{67.4} \\
\hline
\textcolor{gray}{MMV$^\ddagger$~\cite{alayrac2020self}} & \textcolor{gray}{AS+HT (16y)}  & \textcolor{gray}{VAT} & \textcolor{gray}{95.2} & \textcolor{gray}{75.0} \\
\hline
CVRL$^\ddagger$(\footnotesize{R3D-50}) & K400 (28d) & V & \textbf{92.2} & \textbf{66.7}\\
CVRL$^\dagger$(\footnotesize{R3D-50}) & K400 (28d) & V & \textbf{92.9} & \textbf{67.9}\\
CVRL$^\ddagger$(\footnotesize{R3D-50}) & K600 (44d) & V & \textbf{93.4} & \textbf{68.0}\\
CVRL$^\dagger$(\footnotesize{R3D-50}) & K600 (44d) & V & \textbf{93.6} & \textbf{69.4}\\
CVRL$^\ddagger$(\footnotesize{R3D-152(2$\times$)}) & K600 (44d) & V & \textbf{93.9} & \textbf{69.9}\\
CVRL$^\dagger$(\footnotesize{R3D-152(2$\times$)}) & K600 (44d) & V & \textbf{94.4} & \textbf{70.6}\\
\hline
\multicolumn{5}{c}{\multirow{2}{*}{\small{\textbf{Linear Evaluation}}}} \\
\multicolumn{5}{c}{\multirow{2}{*}{}} \\
\hline
\textcolor{gray}{MemDPC$^\mathsection$~\cite{han2020memory}} & \textcolor{gray}{K400 (28d)}  & \textcolor{gray}{VF} & \textcolor{gray}{54.1} & \textcolor{gray}{30.5} \\
\textcolor{gray}{CoCLR$^\dagger$~\cite{han2020coclr}} & \textcolor{gray}{K400 (28d)}  & \textcolor{gray}{VF} & \textcolor{gray}{77.8} & \textcolor{gray}{52.4} \\
\hline
\textcolor{gray}{MIL-NCE$^\ddagger$~\cite{miech2020end}} & \textcolor{gray}{HT (15y)}  & \textcolor{gray}{VT} & \textcolor{gray}{83.4} & \textcolor{gray}{54.8} \\
\hline
\textcolor{gray}{XDC$^\ddagger$~\cite{alwassel2019self}} & \textcolor{gray}{AS (240d)}  & \textcolor{gray}{VA} & \textcolor{gray}{85.3} & \textcolor{gray}{56.0} \\
\textcolor{gray}{MMV-VA$^\ddagger$~\cite{alayrac2020self}} & \textcolor{gray}{AS+HT (16y)}  & \textcolor{gray}{VA} & \textcolor{gray}{86.2} & \textcolor{gray}{61.5} \\
\hline
\textcolor{gray}{Elo$^\mathsection$~\cite{Piergiovanni_2020_CVPR}} & \textcolor{gray}{YT8M (13y)}  & \textcolor{gray}{VFA} & \textcolor{gray}{-} & \textcolor{gray}{64.5} \\
\hline
\textcolor{gray}{MMV$^\ddagger$~\cite{alayrac2020self}} & \textcolor{gray}{AS+HT (16y)}  & \textcolor{gray}{VAT} & \textcolor{gray}{91.8} & \textcolor{gray}{67.1} \\
\hline
CVRL$^\ddagger$(\footnotesize{R3D-50}) & K400 (28d) & V & \textbf{89.2} & \textbf{57.3}\\
CVRL$^\dagger$(\footnotesize{R3D-50}) & K400 (28d) & V & \textbf{89.8} & \textbf{58.3}\\
CVRL$^\ddagger$(\footnotesize{R3D-50}) & K600 (44d) & V & \textbf{90.6} & \textbf{59.7}\\
CVRL$^\dagger$(\footnotesize{R3D-50}) & K600 (44d) & V & \textbf{90.8} & \textbf{59.7}\\
\shline
\end{tabular}
\vspace{2mm}
\caption{\footnotesize{\textbf{Downstream action classification results on UCF-101 and HMDB-51.} 
CVRL shows competitive performance compared with single and muitl-modal methods, by using only the \textbf{V}ision modality on K400 and K600.
$^\dagger$ indicates split-1 accuracy, $^\ddagger$ indicates averaged accuracy on 3 splits, $^\mathsection$ indicates evaluation split(s) not mentioned in paper.}} 
\label{tab:ucf}
\end{table}

\begin{table}[t]
\small
\centering
\begin{tabular}{c|c|c|c|c|c}
\shline
\multirow{2}{*}{Method} & \multirow{2}{*}{\small{Rand.}} & \small{ImageNet} & \small{SimCLR} & \multirow{2}{*}{\small{CVRL}} & \multirow{2}{*}{\textcolor{gray}{\small{Sup.}}} \\
& & \small{infla.} & \small{infla.} & & \\
\hline
mAP & 6.9 & 14.0 & 14.2 & \textbf{16.3} & \textcolor{gray}{19.1}\\
\shline
\end{tabular}
\vspace{2mm}
\caption{\textbf{Downstream action detection results on AVA.} We report mean Average-Precision (mAP) to assess the performance. CVRL outperforms ImageNet inflated and SimCLR inflated. All methods use R3D-50 as the backbone. }
\label{tab:ava}
\end{table}

\begin{table*}[t]
\small
    \begin{minipage}{0.32\linewidth}
        \centering
        \begin{tabular}{c|p{10mm}<{\centering}|cc}
        \shline
        \multirow{2}{*}{Backbone} & Hidden & \multicolumn{2}{c}{Accuracy (\%)} \\
         & layers & top-1 & top-5 \\
        \shline
        \multirow{4}{*}{R3D-50} & 0 & 54.7 & 79.1\\
         & 1 & 62.5 & 84.5\\
         & 2 & 63.0 & 84.8\\
         & 3 & \textbf{63.8} & \textbf{85.2}\\
        \shline
        \end{tabular}
        \vspace{2mm}
        \caption{\textbf{Ablation on hidden layers.}}
        \label{tab:mlp}
    \end{minipage} \hfill
    \begin{minipage}{0.32\linewidth}
      \centering
        \vspace{-0.65mm}
        \begin{tabular}{c|p{10mm}<{\centering}|cc}
        \shline
        \multirow{2}{*}{Backbone} & Batch & \multicolumn{2}{c}{Accuracy (\%)} \\
         & size & top-1 & top-5 \\
        \shline
        \multirow{4}{*}{R3D-50} & 256 & 60.2 & 82.5 \\
         & 512 & 62.9 & 84.7 \\
         & 1024 & \textbf{63.8} & \textbf{85.2} \\
         & 2048 & 61.8 & 84.2 \\
        \shline
        \end{tabular}
        \vspace{2mm}
        \caption{\textbf{Ablation on batch size.}}
        \label{tab:batch_size}
    \end{minipage} \hfill
    \begin{minipage}{0.32\linewidth}
        \centering
        \begin{tabular}{c|p{10mm}<{\centering}|cc}
        \shline
        \multirow{2}{*}{Backbone} & Pretrain & \multicolumn{2}{c}{Accuracy (\%)} \\
         & epochs & top-1 & top-5 \\
        \shline
        \multirow{4}{*}{R3D-50} & 100 & 58.8 & 81.8 \\
         & 200 & 63.8 & 85.2  \\
         & 500 & 65.6 & 86.5  \\
         & 800 & \textbf{66.1} & \textbf{86.8}  \\
        \shline
        \end{tabular}
        \vspace{2mm}
        \caption{\textbf{Ablation on pre-training epochs.}}
        \label{tab:epochs}
    \end{minipage}
  \vspace{-2mm}
  \end{table*}

\vspace{2mm}
\subsection{Ablation Study}
We conduct extensive ablation studies for CVRL based on 200 epochs pre-training on Kinetics-400 and report top-1 linear evaluation accuracy on Kinetics-400.

\paragraph{Temporal interval sampling distribution.} As shown in Figure~\ref{fig:sampling_dis}, we experiment with monotonically decreasing distributions~(a-c), uniform distribution~(d) and monotonically increasing distributions~(e-f). We find decreasing distributions are better. We also compare different power functions for decreasing distribution and choose a simple exponent of 1 (\ie, linear) due to its simplicity and best performance.

\paragraph{Spatial and temporal augmentation.}
We conduct an ablation study on the proposed temporally consistent spatial augmentation.
From results in Table~\ref{tab:ablation}, we have three major observations.
First, both temporal and spatial augmentations are indispensable.
Specifically, using both temporal and spatial augmentations yields 52.3\% top-1 accuracy, significantly outperforming the same model pre-trained with temporal augmentation only (33.0\%) or spatial augmentation only (40.9\%).
Second, the proposed temporally consistent module plays a critical role in achieving good performance. Adding temporal consistency further improves the top-1 accuracy to 63.8\% by a large margin of 11.5\% over 52.3\%. Third, spatial augmentations, which are ignored to some degree in existing self-supervised video representation learning literature, not only matter, but also contribute more than the temporal augmentations. 

\begin{table}[t]
\small
\centering
\setlength{\tabcolsep}{0.6mm}{
\begin{tabular}{ccc|cc}
\shline
\multirow{2}{*}{\begin{tabular}{c}Temporal \\augmentation\end{tabular}} & \multirow{2}{*}{\begin{tabular}{c}Spatial \\augmentation\end{tabular}} & \multirow{2}{*}{\begin{tabular}{c}Temporal\\consistency\end{tabular}} & \multicolumn{2}{c}{Accuracy (\%)} \\
 & & & top-1 & top-5 \\
\shline
\cmark & & & 33.0 & 57.3 \\
& \cmark & & 40.9 & 66.6\\
\cmark & \cmark & & 52.3 & 76.0 \\
\cmark & \cmark & \cmark & \textbf{63.8} & \textbf{85.2}\\
\shline
\end{tabular}}
\vspace{2mm}
\caption{\textbf{Ablation study on data augmentation.}}
\label{tab:ablation}
\vspace{-2mm}
\end{table}

\begin{figure}[t]
\centering
\includegraphics[width=0.95\columnwidth]{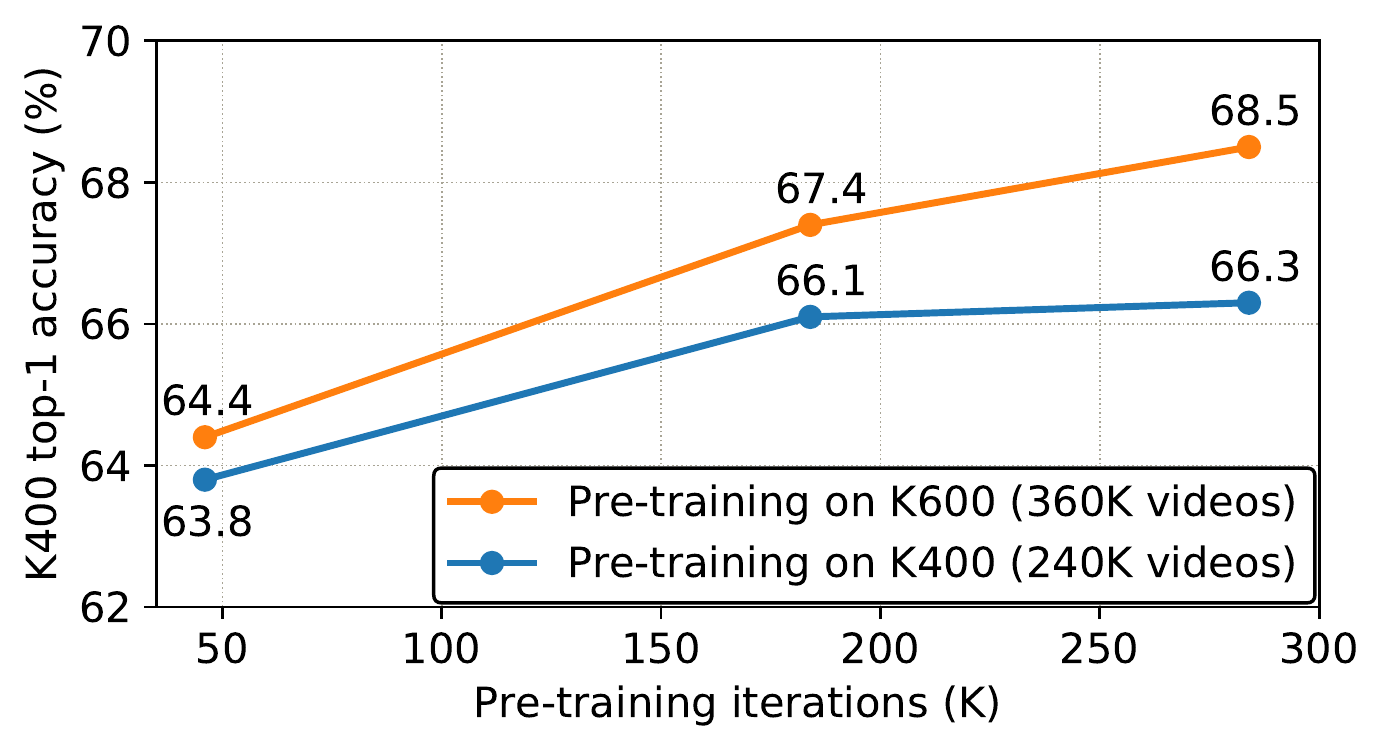}
\caption{\textbf{More data is beneficial for CVRL.} All models are evaluated on the validation set of K400 to provide a fair comparison.}
\label{fig:scala}
\vspace{-2mm}
\end{figure}

\paragraph{More training data.} We study whether using more data would improve the performance of CVRL. We design an evaluation protocol by first pre-training models on different amount of data~(K600 and K400) with same iterations to remove the advantage brought by longer training, and then comparing the performance on same validation set~(K400 val). We verify that the training data of K600 has no overlapping video ids with the validation set of K400. We present results of 46k (200 K400 epochs), 184k (800 K400 epochs) and 284k (800 K600 epochs) pre-training iterations in Figure~\ref{fig:scala}. We find more training data in K600 is beneficial, demonstrating the potential of CVRL's scalability on larger unlabeled datasets.

\paragraph{Layers of projection head.} 
We experiment with different number of hidden layers. Unlike~\cite{chen2020big}, we only use different layers in pre-training and perform the linear evaluation on top of the same backbone by removing the entire projection head. In Table~\ref{tab:mlp}, we can see using 3 hidden layers yields the best performance and we choose this as our default setting.

\paragraph{Batch size.} 
The batch size determines how many negative pairs we use for each positive pair during training. Our experimental results show that a batch size of 1024 already achieves high performance. Larger batch sizes could negatively impact the performance as shown in Table~\ref{tab:batch_size}. 

\paragraph{Pre-training epoch.} As presented in Table~\ref{tab:epochs}, we experiment with pre-training epochs varying from 100 to 800 and find consistent improvement with longer pre-training epochs. We choose 800 epochs as our default setting.

\section{Conclusion}
This work presents a Contrastive Video Representation Learning (CVRL) framework leveraging spatial and temporal cues to learn spatiotemporal representations from unlabeled videos. Extensive studies on linear evaluation, semi-supervised learning and various downstream tasks demonstrate promising results of CVRL. 
In the future, we plan to apply CVRL to a large set of unlabeled videos and incorporate additional modalities into our framework.

\vspace{2mm}
\par\noindent\textbf{Acknowledgment.} We would like to thank Yeqing Li and the TensorFlow TPU team for their infrastructure support; Tsung-Yi Lin, Ting Chen and Yonglong Tian for their valuable feedback.

\clearpage
{\small
\bibliographystyle{ieee_fullname}
\bibliography{ref.bib}

\begin{thebibliography}{10}\itemsep=-1pt

\bibitem{abu2016youtube}
Sami Abu-El-Haija, Nisarg Kothari, Joonseok Lee, Paul Natsev, George Toderici,
  Balakrishnan Varadarajan, and Sudheendra Vijayanarasimhan.
\newblock Youtube-8m: A large-scale video classification benchmark.
\newblock {\em arXiv preprint arXiv:1609.08675}, 2016.

\bibitem{alayrac2020self}
Jean-Baptiste Alayrac, Adri{\`a} Recasens, Rosalia Schneider, Relja
  Arandjelovi{\'c}, Jason Ramapuram, Jeffrey De~Fauw, Lucas Smaira, Sander
  Dieleman, and Andrew Zisserman.
\newblock Self-supervised multimodal versatile networks.
\newblock In {\em NeurIPS}, 2020.

\bibitem{alwassel2019self}
Humam Alwassel, Dhruv Mahajan, Lorenzo Torresani, Bernard Ghanem, and Du Tran.
\newblock Self-supervised learning by cross-modal audio-video clustering.
\newblock In {\em NeurIPS}, 2020.

\bibitem{asano2020labelling}
Yuki~M Asano, Mandela Patrick, Christian Rupprecht, and Andrea Vedaldi.
\newblock Labelling unlabelled videos from scratch with multi-modal
  self-supervision.
\newblock In {\em NeurIPS}, 2020.

\bibitem{bai2020can}
Yutong Bai, Haoqi Fan, Ishan Misra, Ganesh Venkatesh, Yongyi Lu, Yuyin Zhou,
  Qihang Yu, Vikas Chandra, and Alan Yuille.
\newblock Can temporal information help with contrastive self-supervised
  learning?
\newblock {\em arXiv preprint arXiv:2011.13046}, 2020.

\bibitem{benaim2020speednet}
Sagie Benaim, Ariel Ephrat, Oran Lang, Inbar Mosseri, William~T Freeman,
  Michael Rubinstein, Michal Irani, and Tali Dekel.
\newblock Speednet: Learning the speediness in videos.
\newblock In {\em CVPR}, 2020.

\bibitem{swav}
Mathilde Caron, Ishan Misra, Julien Mairal, Priya Goyal, Piotr Bojanowski, and
  Armand Joulin.
\newblock Unsupervised learning of visual features by contrasting cluster
  assignments.
\newblock In {\em NeurIPS}, 2020.

\bibitem{kinetics600}
Joao Carreira, Eric Noland, Andras Banki-Horvath, Chloe Hillier, and Andrew
  Zisserman.
\newblock A short note about kinetics-600.
\newblock {\em arXiv preprint arXiv:1808.01340}, 2018.

\bibitem{i3d}
Joao Carreira and Andrew Zisserman.
\newblock Quo vadis, action recognition? a new model and the kinetics dataset.
\newblock In {\em CVPR}, 2017.

\bibitem{simclr}
Ting Chen, Simon Kornblith, Mohammad Norouzi, and Geoffrey Hinton.
\newblock A simple framework for contrastive learning of visual
  representations.
\newblock In {\em ICML}, 2020.

\bibitem{chen2020big}
Ting Chen, Simon Kornblith, Kevin Swersky, Mohammad Norouzi, and Geoffrey
  Hinton.
\newblock Big self-supervised models are strong semi-supervised learners.
\newblock {\em NeurIPS}, 2020.

\bibitem{randaugment}
Ekin~D Cubuk, Barret Zoph, Jonathon Shlens, and Quoc~V Le.
\newblock Randaugment: Practical automated data augmentation with a reduced
  search space.
\newblock In {\em NeurIPS}, 2020.

\bibitem{diba2019dynamonet}
Ali Diba, Vivek Sharma, Luc~Van Gool, and Rainer Stiefelhagen.
\newblock Dynamonet: Dynamic action and motion network.
\newblock In {\em ICCV}, 2019.

\bibitem{doersch2015unsupervised}
Carl Doersch, Abhinav Gupta, and Alexei~A Efros.
\newblock Unsupervised visual representation learning by context prediction.
\newblock In {\em ICCV}, 2015.

\bibitem{dwibedi2019temporal}
Debidatta Dwibedi, Yusuf Aytar, Jonathan Tompson, Pierre Sermanet, and Andrew
  Zisserman.
\newblock Temporal cycle-consistency learning.
\newblock In {\em CVPR}, 2019.

\bibitem{x3d}
Christoph Feichtenhofer.
\newblock X3d: Expanding architectures for efficient video recognition.
\newblock In {\em CVPR}, 2020.

\bibitem{slowfast}
Christoph Feichtenhofer, Haoqi Fan, Jitendra Malik, and Kaiming He.
\newblock Slowfast networks for video recognition.
\newblock In {\em ICCV}, 2019.

\bibitem{fernando2017self}
Basura Fernando, Hakan Bilen, Efstratios Gavves, and Stephen Gould.
\newblock Self-supervised video representation learning with odd-one-out
  networks.
\newblock In {\em CVPR}, 2017.

\bibitem{gan2018geometry}
Chuang Gan, Boqing Gong, Kun Liu, Hao Su, and Leonidas~J Guibas.
\newblock Geometry guided convolutional neural networks for self-supervised
  video representation learning.
\newblock In {\em CVPR}, 2018.

\bibitem{gemmeke2017audio}
Jort~F Gemmeke, Daniel~PW Ellis, Dylan Freedman, Aren Jansen, Wade Lawrence,
  R~Channing Moore, Manoj Plakal, and Marvin Ritter.
\newblock Audio set: An ontology and human-labeled dataset for audio events.
\newblock In {\em ICASSP}, 2017.

\bibitem{ghadiyaram2019large}
Deepti Ghadiyaram, Du Tran, and Dhruv Mahajan.
\newblock Large-scale weakly-supervised pre-training for video action
  recognition.
\newblock In {\em CVPR}, 2019.

\bibitem{gidaris2018unsupervised}
Spyros Gidaris, Praveer Singh, and Nikos Komodakis.
\newblock Unsupervised representation learning by predicting image rotations.
\newblock In {\em ICLR}, 2018.

\bibitem{girdhar2019distinit}
Rohit Girdhar, Du Tran, Lorenzo Torresani, and Deva Ramanan.
\newblock Distinit: Learning video representations without a single labeled
  video.
\newblock In {\em ICCV}, 2019.

\bibitem{gordon2020watching}
Daniel Gordon, Kiana Ehsani, Dieter Fox, and Ali Farhadi.
\newblock Watching the world go by: Representation learning from unlabeled
  videos.
\newblock {\em arXiv preprint arXiv:2003.07990}, 2020.

\bibitem{goyal2017accurate}
Priya Goyal, Piotr Doll{\'a}r, Ross Girshick, Pieter Noordhuis, Lukasz
  Wesolowski, Aapo Kyrola, Andrew Tulloch, Yangqing Jia, and Kaiming He.
\newblock Accurate, large minibatch sgd: Training imagenet in 1 hour.
\newblock {\em arXiv preprint arXiv:1706.02677}, 2017.

\bibitem{byol}
Jean-Bastien Grill, Florian Strub, Florent Altch{\'e}, Corentin Tallec,
  Pierre~H Richemond, Elena Buchatskaya, Carl Doersch, Bernardo~Avila Pires,
  Zhaohan~Daniel Guo, Mohammad~Gheshlaghi Azar, et~al.
\newblock Bootstrap your own latent: A new approach to self-supervised
  learning.
\newblock In {\em NeurIPS}, 2020.

\bibitem{gu2018ava}
Chunhui Gu, Chen Sun, David~A Ross, Carl Vondrick, Caroline Pantofaru, Yeqing
  Li, Sudheendra Vijayanarasimhan, George Toderici, Susanna Ricco, Rahul
  Sukthankar, et~al.
\newblock Ava: A video dataset of spatio-temporally localized atomic visual
  actions.
\newblock In {\em CVPR}, 2018.

\bibitem{han2019video}
Tengda Han, Weidi Xie, and Andrew Zisserman.
\newblock Video representation learning by dense predictive coding.
\newblock In {\em ICCV Workshops}, 2019.

\bibitem{han2020memory}
Tengda Han, Weidi Xie, and Andrew Zisserman.
\newblock Memory-augmented dense predictive coding for video representation
  learning.
\newblock In {\em ECCV}, 2020.

\bibitem{han2020coclr}
Tengda Han, Weidi Xie, and Andrew Zisserman.
\newblock Self-supervised co-training for video representation learning.
\newblock In {\em NeurIPS}, 2020.

\bibitem{resnet3d}
Kensho Hara, Hirokatsu Kataoka, and Yutaka Satoh.
\newblock Can spatiotemporal 3d cnns retrace the history of 2d cnns and
  imagenet?
\newblock In {\em CVPR}, 2018.

\bibitem{moco}
Kaiming He, Haoqi Fan, Yuxin Wu, Saining Xie, and Ross Girshick.
\newblock Momentum contrast for unsupervised visual representation learning.
\newblock In {\em CVPR}, 2020.

\bibitem{resnet}
Kaiming He, Xiangyu Zhang, Shaoqing Ren, and Jian Sun.
\newblock Deep residual learning for image recognition.
\newblock In {\em CVPR}, 2016.

\bibitem{he2019bag}
Tong He, Zhi Zhang, Hang Zhang, Zhongyue Zhang, Junyuan Xie, and Mu Li.
\newblock Bag of tricks for image classification with convolutional neural
  networks.
\newblock In {\em CVPR}, 2019.

\bibitem{henaff2019data}
Olivier~J H{\'e}naff, Aravind Srinivas, Jeffrey De~Fauw, Ali Razavi, Carl
  Doersch, SM Eslami, and Aaron van~den Oord.
\newblock Data-efficient image recognition with contrastive predictive coding.
\newblock {\em arXiv preprint arXiv:1905.09272}, 2019.

\bibitem{mobilenets}
Andrew~G Howard, Menglong Zhu, Bo Chen, Dmitry Kalenichenko, Weijun Wang,
  Tobias Weyand, Marco Andreetto, and Hartwig Adam.
\newblock Mobilenets: Efficient convolutional neural networks for mobile vision
  applications.
\newblock {\em arXiv preprint arXiv:1704.04861}, 2017.

\bibitem{jing2018self}
Longlong Jing and Yingli Tian.
\newblock Self-supervised spatiotemporal feature learning by video geometric
  transformations.
\newblock {\em arXiv preprint arXiv:1811.11387}, 2018.

\bibitem{kay2017kinetics}
Will Kay, Joao Carreira, Karen Simonyan, Brian Zhang, Chloe Hillier, Sudheendra
  Vijayanarasimhan, Fabio Viola, Tim Green, Trevor Back, Paul Natsev, et~al.
\newblock The kinetics human action video dataset.
\newblock {\em arXiv preprint arXiv:1705.06950}, 2017.

\bibitem{kim2019self}
Dahun Kim, Donghyeon Cho, and In~So Kweon.
\newblock Self-supervised video representation learning with space-time cubic
  puzzles.
\newblock In {\em AAAI}, 2019.

\bibitem{korbar2018cooperative}
Bruno Korbar, Du Tran, and Lorenzo Torresani.
\newblock Cooperative learning of audio and video models from self-supervised
  synchronization.
\newblock In {\em NeurIPS}, 2018.

\bibitem{kuehne2011hmdb}
Hildegard Kuehne, Hueihan Jhuang, Est{\'\i}baliz Garrote, Tomaso Poggio, and
  Thomas Serre.
\newblock Hmdb: a large video database for human motion recognition.
\newblock In {\em ICCV}, 2011.

\bibitem{lee2017unsupervised}
Hsin-Ying Lee, Jia-Bin Huang, Maneesh Singh, and Ming-Hsuan Yang.
\newblock Unsupervised representation learning by sorting sequences.
\newblock In {\em ICCV}, 2017.

\bibitem{lotter2016deep}
William Lotter, Gabriel Kreiman, and David Cox.
\newblock Deep predictive coding networks for video prediction and unsupervised
  learning.
\newblock {\em arXiv preprint arXiv:1605.08104}, 2016.

\bibitem{sift}
David~G Lowe.
\newblock Distinctive image features from scale-invariant keypoints.
\newblock {\em IJCV}, 2004.

\bibitem{miech2020end}
Antoine Miech, Jean-Baptiste Alayrac, Lucas Smaira, Ivan Laptev, Josef Sivic,
  and Andrew Zisserman.
\newblock End-to-end learning of visual representations from uncurated
  instructional videos.
\newblock In {\em CVPR}, 2020.

\bibitem{miech2019howto100m}
Antoine Miech, Dimitri Zhukov, Jean-Baptiste Alayrac, Makarand Tapaswi, Ivan
  Laptev, and Josef Sivic.
\newblock Howto100m: Learning a text-video embedding by watching hundred
  million narrated video clips.
\newblock In {\em ICCV}, 2019.

\bibitem{noroozi2016unsupervised}
Mehdi Noroozi and Paolo Favaro.
\newblock Unsupervised learning of visual representations by solving jigsaw
  puzzles.
\newblock In {\em ECCV}, 2016.

\bibitem{cpc}
Aaron van~den Oord, Yazhe Li, and Oriol Vinyals.
\newblock Representation learning with contrastive predictive coding.
\newblock {\em arXiv preprint arXiv:1807.03748}, 2018.

\bibitem{pathak2017learning}
Deepak Pathak, Ross Girshick, Piotr Doll{\'a}r, Trevor Darrell, and Bharath
  Hariharan.
\newblock Learning features by watching objects move.
\newblock In {\em CVPR}, 2017.

\bibitem{pathak2016context}
Deepak Pathak, Philipp Krahenbuhl, Jeff Donahue, Trevor Darrell, and Alexei~A
  Efros.
\newblock Context encoders: Feature learning by inpainting.
\newblock In {\em CVPR}, 2016.

\bibitem{patrick2020multi}
Mandela Patrick, Yuki~M Asano, Ruth Fong, Jo{\~a}o~F Henriques, Geoffrey Zweig,
  and Andrea Vedaldi.
\newblock Multi-modal self-supervision from generalized data transformations.
\newblock {\em arXiv preprint arXiv:2003.04298}, 2020.

\bibitem{Piergiovanni_2020_CVPR}
AJ Piergiovanni, Anelia Angelova, and Michael~S. Ryoo.
\newblock Evolving losses for unsupervised video representation learning.
\newblock In {\em CVPR}, 2020.

\bibitem{Purushwalkam12020Demystifying}
Senthil Purushwalkam1 and Abhinav Gupta.
\newblock Demystifying contrastive self-supervised learning: Invariances,
  augmentations and dataset biases.
\newblock In {\em NeurIPS}, 2020.

\bibitem{ren2015faster}
Shaoqing Ren, Kaiming He, Ross Girshick, and Jian Sun.
\newblock Faster r-cnn: Towards real-time object detection with region proposal
  networks.
\newblock In {\em NeurIPS}, 2015.

\bibitem{3dsift}
Paul Scovanner, Saad Ali, and Mubarak Shah.
\newblock A 3-dimensional sift descriptor and its application to action
  recognition.
\newblock In {\em ACM MM}, 2007.

\bibitem{sermanet2018time}
Pierre Sermanet, Corey Lynch, Yevgen Chebotar, Jasmine Hsu, Eric Jang, Stefan
  Schaal, Sergey Levine, and Google Brain.
\newblock Time-contrastive networks: Self-supervised learning from video.
\newblock In {\em ICRA}, 2018.

\bibitem{ucf101}
Khurram Soomro, Amir~Roshan Zamir, and Mubarak Shah.
\newblock Ucf101: A dataset of 101 human actions classes from videos in the
  wild.
\newblock {\em arXiv preprint arXiv:1212.0402}, 2012.

\bibitem{srivastava2015unsupervised}
Nitish Srivastava, Elman Mansimov, and Ruslan Salakhudinov.
\newblock Unsupervised learning of video representations using lstms.
\newblock In {\em ICML}, 2015.

\bibitem{sun2019learning}
Chen Sun, Fabien Baradel, Kevin Murphy, and Cordelia Schmid.
\newblock Learning video representations using contrastive bidirectional
  transformer.
\newblock {\em arXiv preprint arXiv:1906.05743}, 2019.

\bibitem{sun2019videobert}
Chen Sun, Austin Myers, Carl Vondrick, Kevin Murphy, and Cordelia Schmid.
\newblock Videobert: A joint model for video and language representation
  learning.
\newblock In {\em ICCV}, 2019.

\bibitem{tian2019contrastive}
Yonglong Tian, Dilip Krishnan, and Phillip Isola.
\newblock Contrastive multiview coding.
\newblock In {\em ECCV}, 2020.

\bibitem{tian2020makes}
Yonglong Tian, Chen Sun, Ben Poole, Dilip Krishnan, Cordelia Schmid, and
  Phillip Isola.
\newblock What makes for good views for contrastive learning.
\newblock In {\em NeurIPS}, 2020.

\bibitem{c3d}
Du Tran, Lubomir Bourdev, Rob Fergus, Lorenzo Torresani, and Manohar Paluri.
\newblock Learning spatiotemporal features with 3d convolutional networks.
\newblock In {\em ICCV}, 2015.

\bibitem{vogel2002computational}
Curtis~R Vogel.
\newblock {\em Computational methods for inverse problems}.
\newblock SIAM, 2002.

\bibitem{vondrick2018tracking}
Carl Vondrick, Abhinav Shrivastava, Alireza Fathi, Sergio Guadarrama, and Kevin
  Murphy.
\newblock Tracking emerges by colorizing videos.
\newblock In {\em ECCV}, 2018.

\bibitem{wang2019self}
Jiangliu Wang, Jianbo Jiao, Linchao Bao, Shengfeng He, Yunhui Liu, and Wei Liu.
\newblock Self-supervised spatio-temporal representation learning for videos by
  predicting motion and appearance statistics.
\newblock In {\em CVPR}, 2019.

\bibitem{wang2020self}
Jiangliu Wang, Jianbo Jiao, and Yun-Hui Liu.
\newblock Self-supervised video representation learning by pace prediction.
\newblock In {\em ECCV}, 2020.

\bibitem{wang2015unsupervised}
Xiaolong Wang and Abhinav Gupta.
\newblock Unsupervised learning of visual representations using videos.
\newblock In {\em ICCV}, 2015.

\bibitem{CVPR2019_CycleTime}
Xiaolong Wang, Allan Jabri, and Alexei~A. Efros.
\newblock Learning correspondence from the cycle-consistency of time.
\newblock In {\em CVPR}, 2019.

\bibitem{wu2018unsupervised}
Zhirong Wu, Yuanjun Xiong, Stella~X Yu, and Dahua Lin.
\newblock Unsupervised feature learning via non-parametric instance
  discrimination.
\newblock In {\em CVPR}, 2018.

\bibitem{s3d}
Saining Xie, Chen Sun, Jonathan Huang, Zhuowen Tu, and Kevin Murphy.
\newblock Rethinking spatiotemporal feature learning: Speed-accuracy trade-offs
  in video classification.
\newblock In {\em ECCV}, 2018.

\bibitem{xu2019self}
Dejing Xu, Jun Xiao, Zhou Zhao, Jian Shao, Di Xie, and Yueting Zhuang.
\newblock Self-supervised spatiotemporal learning via video clip order
  prediction.
\newblock In {\em CVPR}, 2019.

\bibitem{yang2020video}
Ceyuan Yang, Yinghao Xu, Bo Dai, and Bolei Zhou.
\newblock Video representation learning with visual tempo consistency.
\newblock {\em arXiv preprint arXiv:2006.15489}, 2020.

\bibitem{yao2020seco}
Ting Yao, Yiheng Zhang, Zhaofan Qiu, Yingwei Pan, and Tao Mei.
\newblock Seco: Exploring sequence supervision for unsupervised representation
  learning.
\newblock {\em arXiv preprint arXiv:2008.00975}, 2020.

\bibitem{Yao_2020_CVPR}
Yuan Yao, Chang Liu, Dezhao Luo, Yu Zhou, and Qixiang Ye.
\newblock Video playback rate perception for self-supervised spatio-temporal
  representation learning.
\newblock In {\em CVPR}, 2020.

\bibitem{ye2019unsupervised}
Mang Ye, Xu Zhang, Pong~C Yuen, and Shih-Fu Chang.
\newblock Unsupervised embedding learning via invariant and spreading instance
  feature.
\newblock In {\em CVPR}, 2019.

\bibitem{zhang2016colorful}
Richard Zhang, Phillip Isola, and Alexei~A Efros.
\newblock Colorful image colorization.
\newblock In {\em ECCV}, 2016.

\bibitem{zhang2017split}
Richard Zhang, Phillip Isola, and Alexei~A Efros.
\newblock Split-brain autoencoders: Unsupervised learning by cross-channel
  prediction.
\newblock In {\em CVPR}, 2017.

\end{thebibliography}
}

\clearpage
\begin{appendix}

\section{Details on Temporal Interval Sampling}
\label{appdix:sampling}

Here we describe how to sample the temporal interval $t \in [0, T]$ from a given distribution $P(t)$. Suppose $P(t)$ is a power function:
\begin{equation}
    P(t) = at^b + c,
\end{equation}
where $a$, $b$ and $c$ are constants.
We adopt the technique of inverse transform sampling~\cite{vogel2002computational} by first calculating the cumulative distribution function (CDF) $F(t)$ of $P(t)$ as:
\begin{equation}
    F(t) = \int_{-\infty}^{t} P(x) \mathop{dx} = \frac{a}{b+1} t^{b+1} + ct,
\end{equation}
where $t \in [0, T]$.
To sample a temporal interval $t$, we then generate a random variable $v \sim U(0, 1)$ from a standard uniform distribution and calculate $t = F^{-1}(v)$. Notice that it is difficult to directly compute the closed-form solution of the inverse function of $F(\cdot)$. Considering the facts that the temporal interval $t$ is an integer representing the number of frames between the start frames of two clips and $F(\cdot)$ is monotonically increasing, we use a simple binary search method in Algorithm~\ref{alg:sample} to find $t$. The algorithm is demonstrated below and the complexity is $\mathcal{O}(\log T)$.

\begin{figure*}[t]
\centering
\subfigure{
\includegraphics[width=0.48\columnwidth]{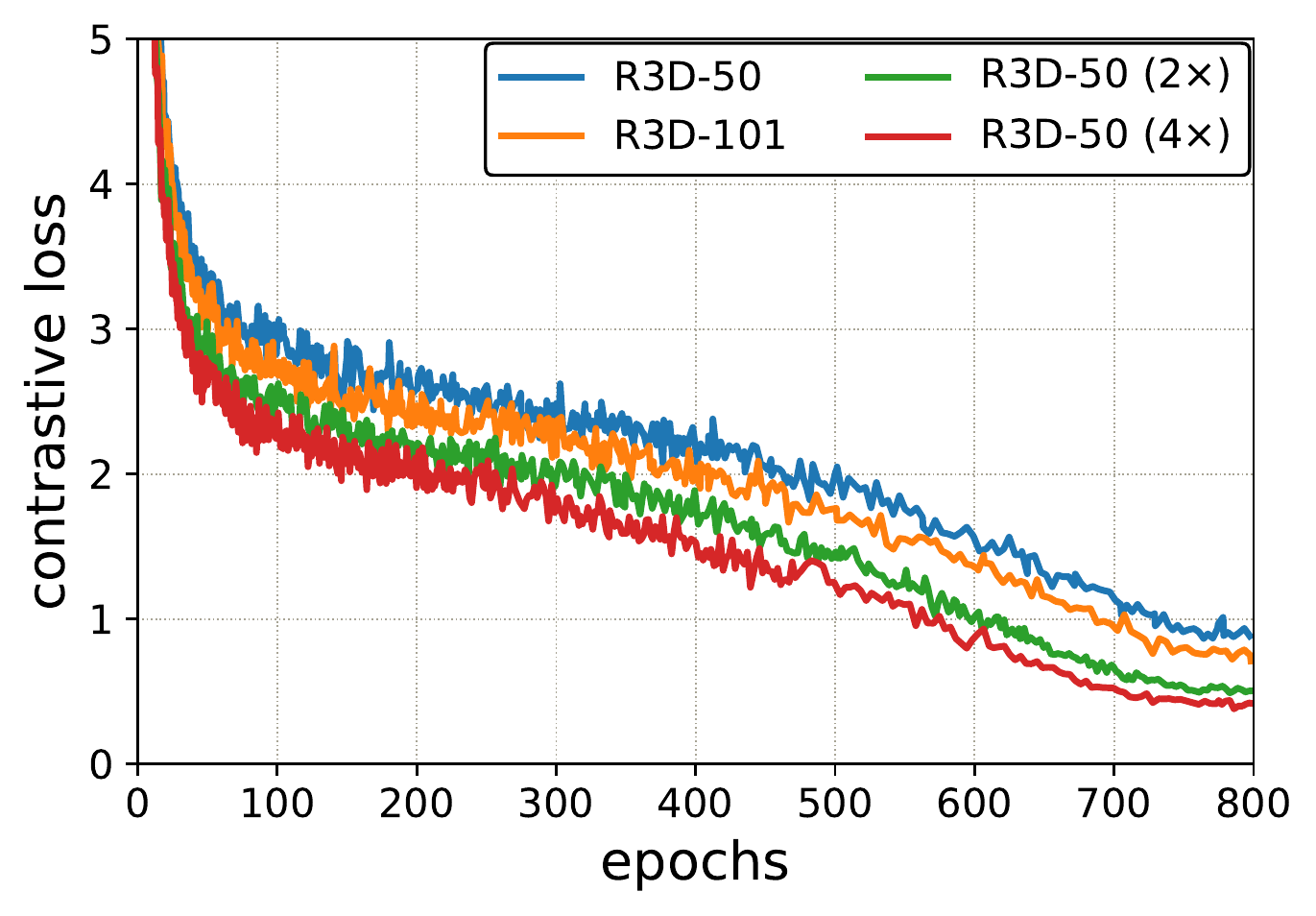}
}
\subfigure{
\includegraphics[width=0.50\columnwidth]{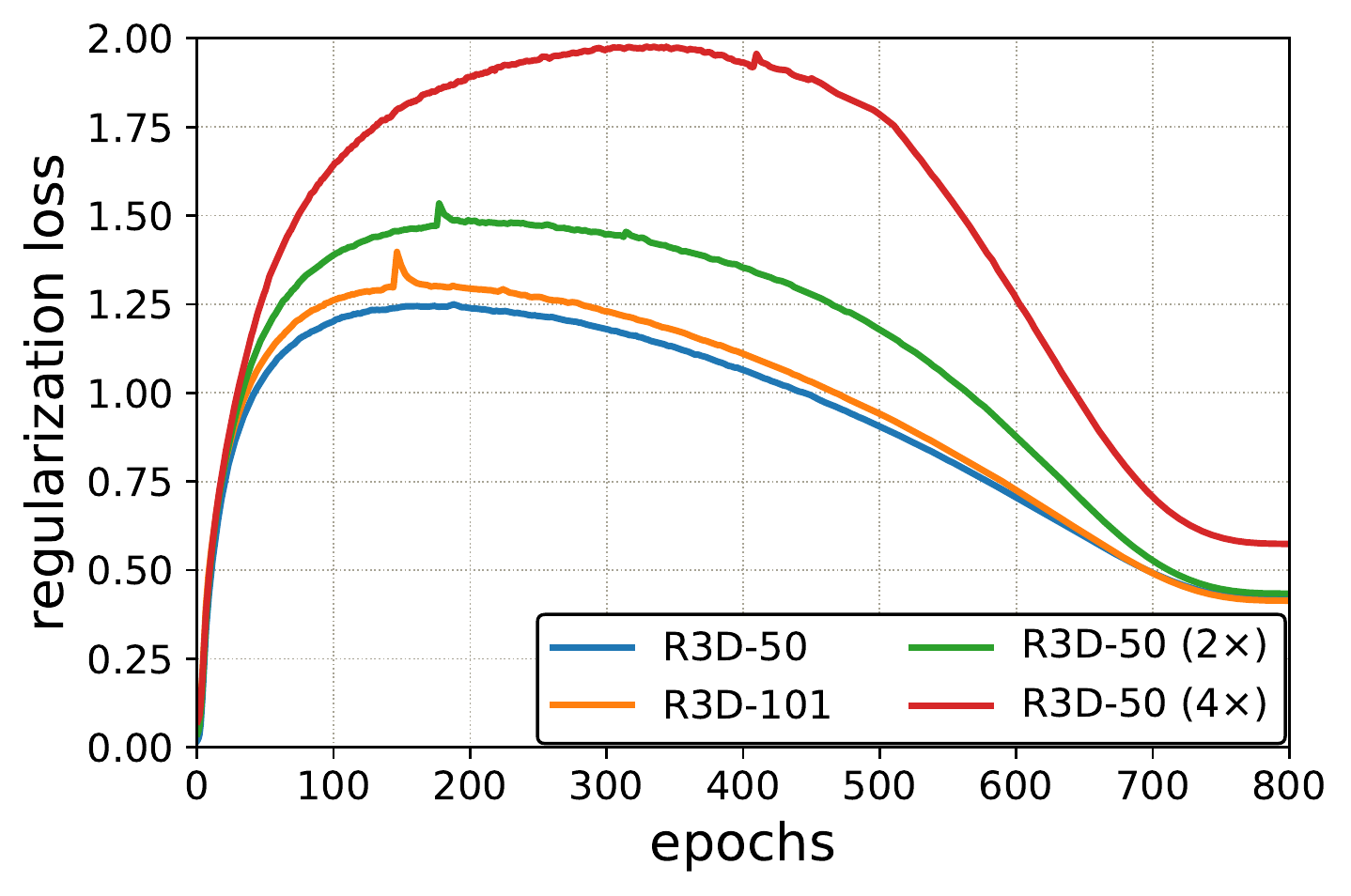}
}
\subfigure{
\includegraphics[width=0.48\columnwidth]{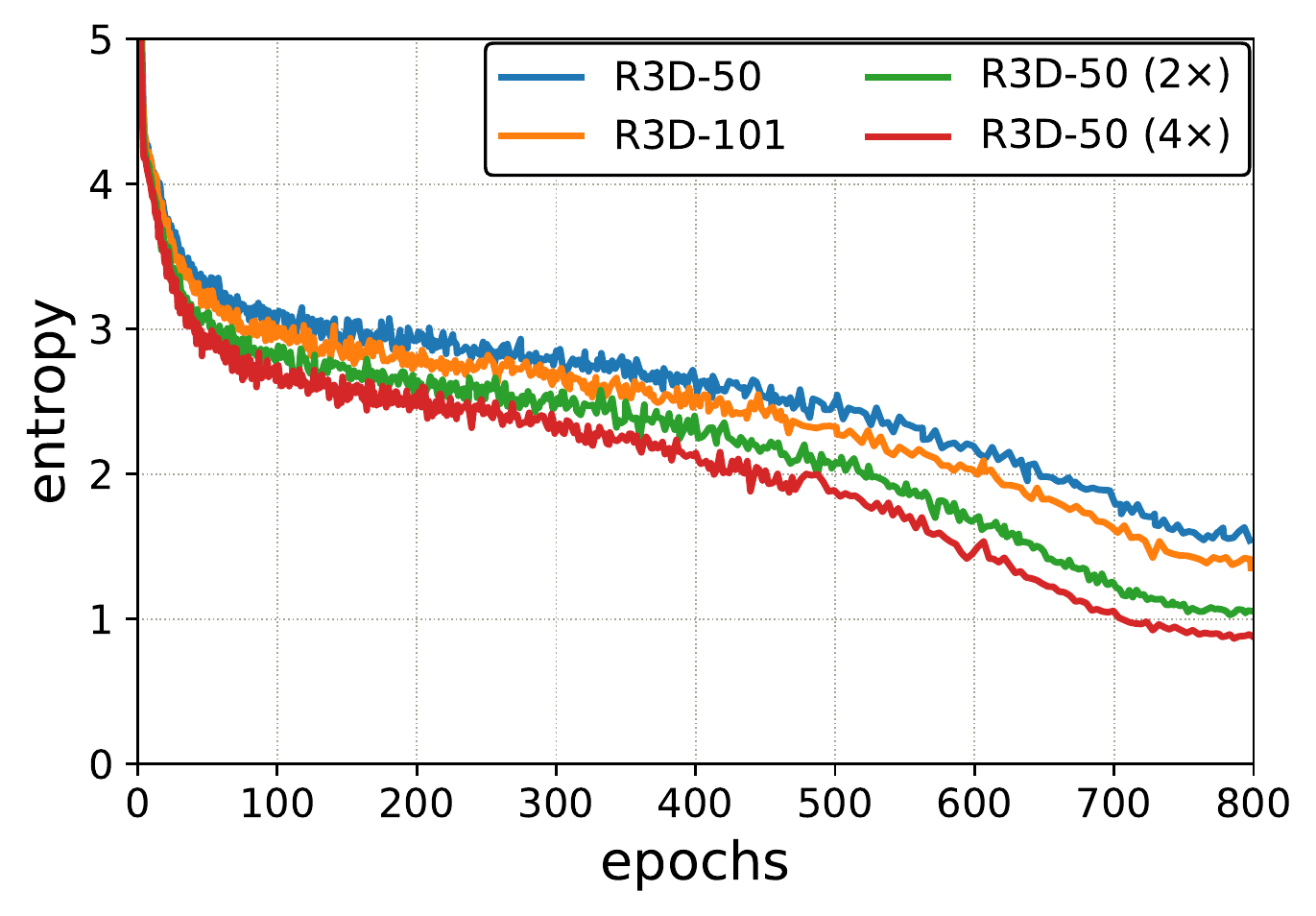}
}
\subfigure{
\includegraphics[width=0.50\columnwidth]{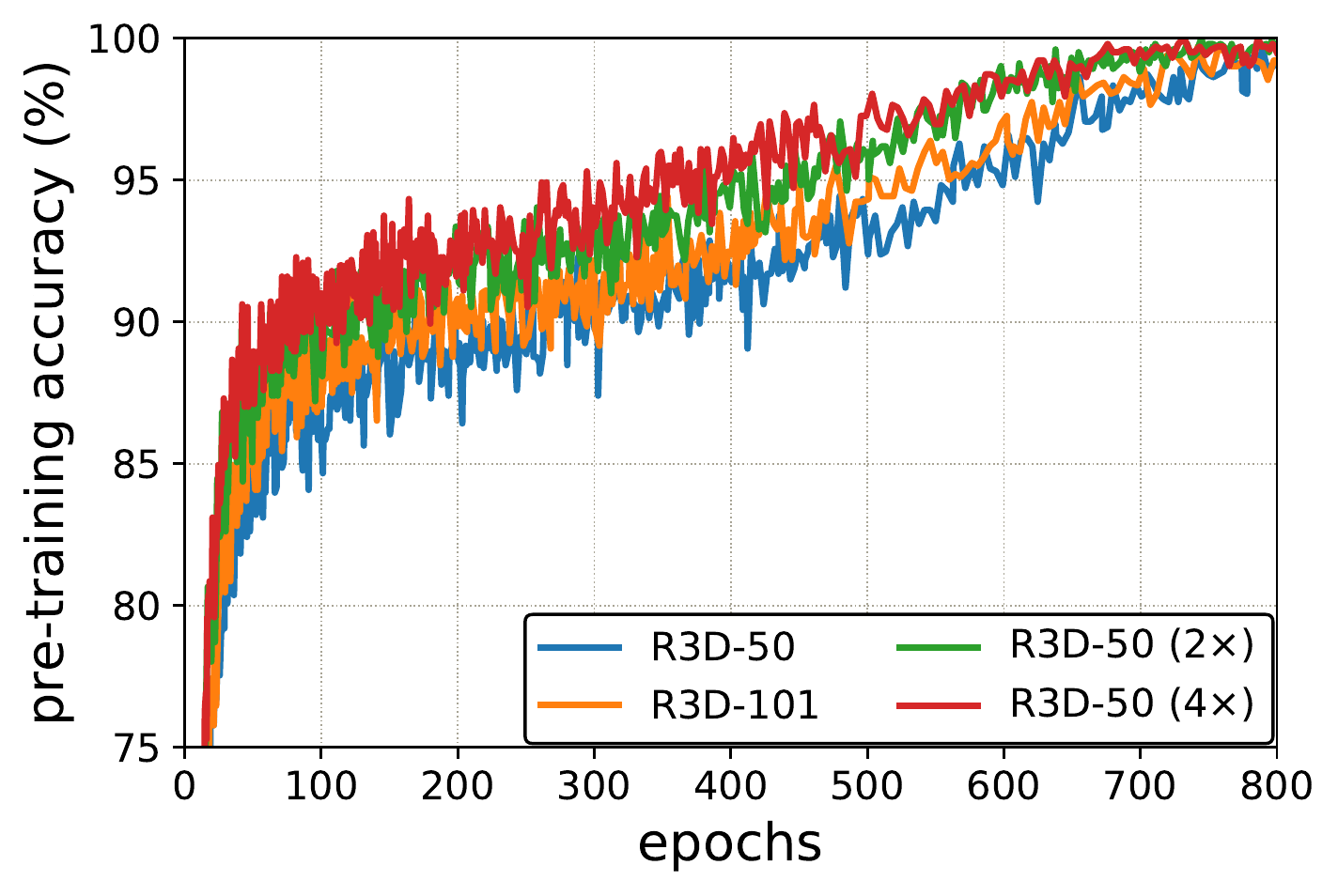}
}
\caption{\textbf{Model pre-training statistics:} contrastive loss, regularization loss, entropy and pre-training accuracy on Kinetics-400.}
\label{fig:train_curves}
\end{figure*}

\begin{algorithm}[ht]
\small
\caption{\label{alg:sample} \small{Temporal Interval Sampling}}
\begin{algorithmic}
\REQUIRE random variable $v \sim U(0, 1)$, CDF function $F(\cdot)$ \\
\STATE $upper\_bound = T$
\STATE $lower\_bound = 0$
\WHILE{$upper\_bound - lower\_bound > 1$}
\STATE $~~~~$ $t = \text{int}((upper\_bound + lower\_bound) / 2)$
\STATE $~~~~$ \textbf{if} $F(t) > v$ \textbf{do}
\STATE $~~~~~~~~~~~~$ $upper\_bound = t$
\STATE $~~~~$ \textbf{else} \textbf{do}
\STATE $~~~~~~~~~~~~$ $lower\_bound = t$ 
\ENDWHILE
\ENSURE temporal interval $t \approx F^{-1}(v)$
\end{algorithmic}
\end{algorithm}

\vspace{1cm}
\section{Additional Results}

\subsection{Semi-Supervised Learning on Kinetics-600}
\label{appdix:semi_k600}
We also conduct semi-supervised learning on K600. Similar to K400, we sample 1\% and 10\% videos from each class in the training set, forming two balanced subsets, respectively. The evaluation set remains the same. As in Table~\ref{tab:semi_k600}, CVRL shows strong performance especially when there is only 1\% labeled data.

\definecolor{upcolor}{RGB}{0,0,0}
\begin{table}[t]
\small
\centering
\begin{tabular}{c|c|ll}
\shline
\multirow{2}{*}{Method} & \multicolumn{1}{c|}{\multirow{2}{*}{Backbone}} & \multicolumn{2}{c}{K600 Top-1 Acc. ($\Delta$ \vs Sup.)} \\
 \cline{3-4}
 & & \multicolumn{1}{c}{1\% label} & \multicolumn{1}{c}{10\% label} \\
 \cline{3-4}
\shline
{\hspace{-3mm}\begin{tabular}{c}Supervised\end{tabular}\hspace{-3mm}} & R3D-50 & ~~4.3 & 45.3 \\
{\hspace{-3mm}\begin{tabular}{c}SimCLR infla.\end{tabular}\hspace{-3mm}} & R3D-50 & 16.9 (\textcolor{upcolor}{12.6$\uparrow$}) & 51.4 (\textcolor{upcolor}{6.1$\uparrow$}) \\
{\hspace{-3mm}\begin{tabular}{c}ImageNet infla.\end{tabular}\hspace{-3mm}} & R3D-50 & 19.7 (\textcolor{upcolor}{15.4$\uparrow$}) & 48.3 (\textcolor{upcolor}{3.0$\uparrow$}) \\
\hline
{CVRL} & R3D-50 & \textbf{36.7} (\textcolor{upcolor}{\textbf{32.4}$\uparrow$})& \textbf{56.1} (\textcolor{upcolor}{\textbf{10.8}$\uparrow$}) \\
\shline
\end{tabular}
\vspace{2mm}
\caption{\textbf{Semi-supervised learning results on Kinetics-600.}}
\label{tab:semi_k600}
\vspace{2mm}
\end{table}

\subsection{Comparison with RandAugment}
We are interested in the performance of strong spatial augmentations that are widely used in supervised learning. We experiment with RandAugment~\cite{randaugment} to randomly select 2 operators from a pool of 14. We conduct experiments with 200 epochs pre-training on Kinetics-400~\cite{kay2017kinetics}. For linear evaluation, RandAugment with temporal consistency achieves 54.2\% top-1 accuracy as shown in Table~\ref{tab:aug}, which is lower than our temporally consistent spatial augmentation presented in Algorithm~\ref{alg:aug}, implying that strong augmentations optimized for supervised image recognition do not necessarily perform as well in the self-supervised video representation learning.

\begin{table}[t]
\small
\centering
\begin{tabular}{c|cc}
\shline
\multirow{2}{*}{Augmentation method} &\multicolumn{2}{c}{Accuracy (\%)} \\
 & top-1 & top-5 \\
\shline
RandAugment w/ temporal consistency & 54.2 & 77.9 \\
Proposed & \textbf{63.8} & \textbf{85.2}\\
\shline
\end{tabular}
\vspace{2mm}
\caption{\textbf{Performance of different spatial augmentations in pre-training (200 epochs).} Our proposed augmentation method outperforms RandAugment with temporal consistency. }
\label{tab:aug}
\end{table}

\section{Illustrations}
\subsection{Pre-Training and Linear Evaluation}
\label{appdix:stats}
\begin{figure}[t]
\centering
\includegraphics[width=0.9\columnwidth]{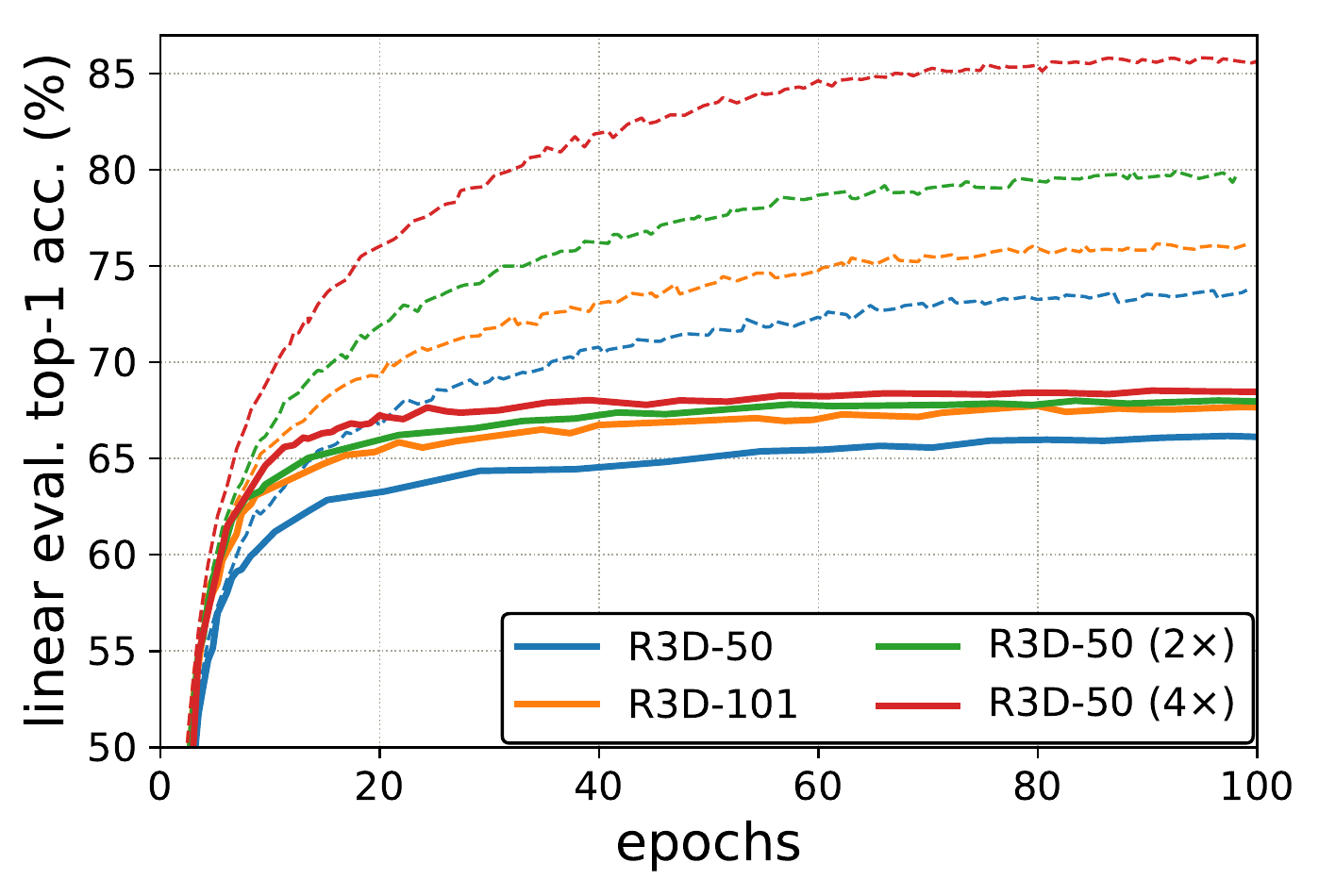}
\caption{\textbf{Linear evaluation training (dashed-line) and evaluation (solid-line) top-1 accuracy} on Kinetics-400.}
\label{fig:finetune_curves}
\vspace{-2mm}
\end{figure}

More detailed pre-training statistics on Kinetics-400~\cite{kay2017kinetics} are illustrated in Figure~\ref{fig:train_curves}. We display four metrics: (1) contrastive loss, (2) regularization loss, (3) entropy and (4) pre-training accuracy. The total loss is the sum of contrastive loss and regularization loss.
We also provide linear evaluation statistics in Figure~\ref{fig:finetune_curves}, where all models are pre-trained on Kinetics-400 for 800 epochs corresponding to Figure~\ref{fig:train_curves}.

\subsection{Temporally Consistent Spatial Augmentation}
\label{appdix:tmp_consis}
\begin{figure}[t]
\centering
\includegraphics[width=0.9\columnwidth]{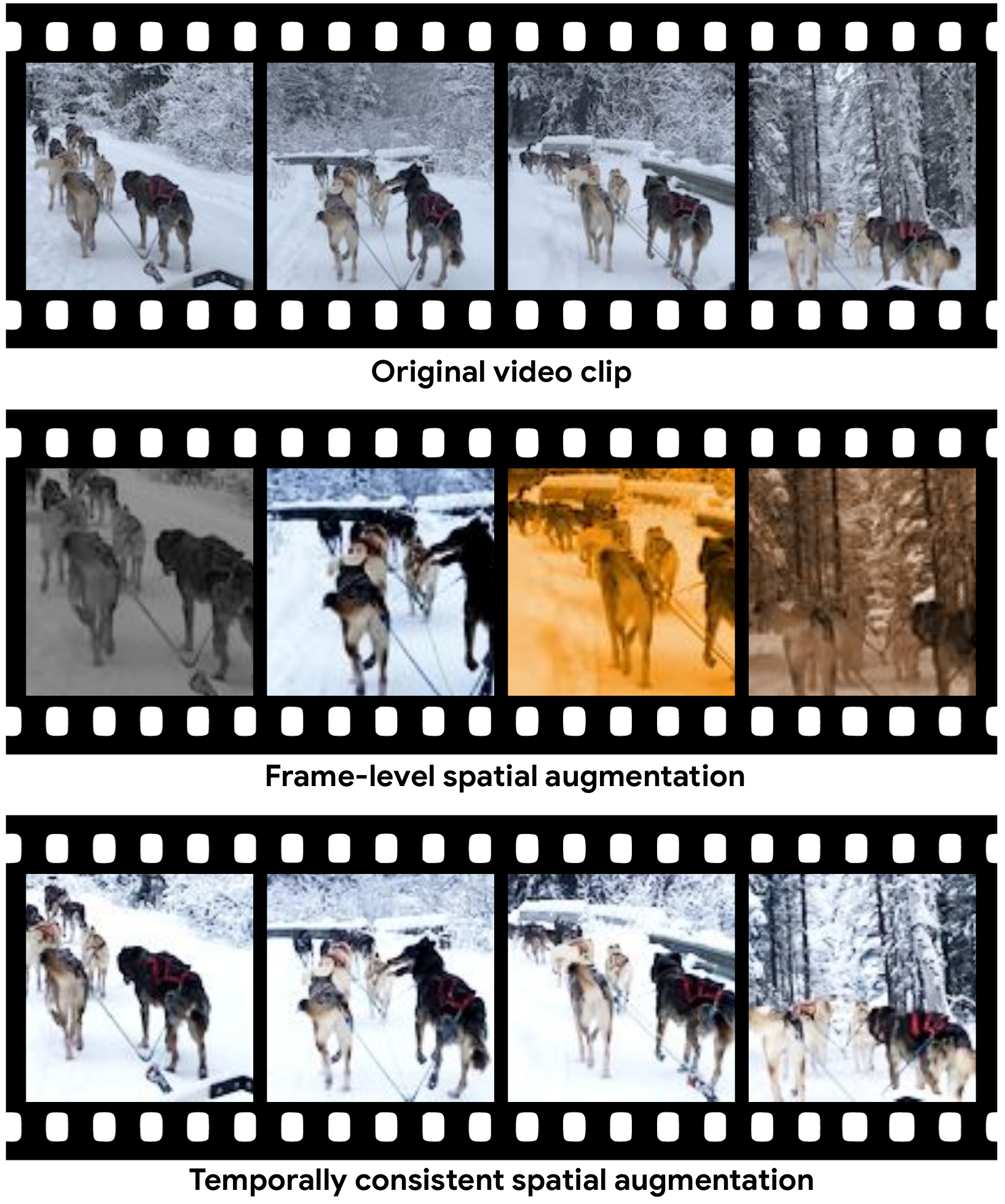}
\caption{\textbf{Illustration of temporally consistent spatial augmentation.} The middle row indicates frame-level spatial augmentations without temporal consistency which would be detrimental to the video representation learning.}
\label{fig:augmentation}
\end{figure}

We illustrate the proposed temporally consistent spatial augmentation method in Figure~\ref{fig:augmentation}. Given an original video clip~(top row), simply applying spatial augmentations to each frame independently would break the motion cues across frames~(middle row). The proposed temporally consistent spatial augmentation~(bottom row) would augment the spatial domain of the video clip while maintaining their natural temporal motion changes. 

\end{appendix}
\end{document}